%% file: main.tex
\documentclass[letterpaper]{article} 
\usepackage[draft]{aaai25}  
\usepackage{times}  
\usepackage{helvet}  
\usepackage{courier}  
\usepackage[hyphens]{url}  
\usepackage{graphicx} 
\usepackage{natbib}  
\usepackage{caption} 
\usepackage{algpseudocode} 
\usepackage{algorithm}

\def\ie{\emph{i.e.}}

\usepackage{newfloat}
\usepackage{listings}
\DeclareCaptionStyle{ruled}{labelfont=normalfont,labelsep=colon,strut=off} 
\floatstyle{ruled}
\newfloat{listing}{tb}{lst}{}
\floatname{listing}{Listing}
\title{
Numerical Pruning for Efficient Autoregressive Models
}
\author{
    Xuan Shen$^1$\thanks{Work done during internship at Adobe Research}, Zhao Song$^2$, Yufa Zhou$^3$, Bo Chen$^4$, Jing Liu$^5$, Ruiyi Zhang$^2$, Ryan A. Rossi$^2$, Hao Tan$^2$, Tong Yu$^2$, Xiang Chen$^2$, Yufan Zhou$^2$, Tong Sun$^2$, Pu Zhao$^1$, Yanzhi Wang$^1$\thanks{Corresponding Author}, Jiuxiang Gu$^2$\footnotemark[2]
}
\affiliations{
    $^1$Northeastern University, $^2$Adobe Research, $^3$University of Pennsylvania, \\ $^4$Middle Tennessee State University, $^5$Monash University

    \{shen.xu, p.zhao, yanz.wang\}@northeastern.edu,  jigu@adobe.com
}

\usepackage{bibentry}

\usepackage{booktabs}
\usepackage{multirow}
\usepackage{multicol}
\usepackage{xcolor}
\usepackage{amsthm}
\usepackage{amsfonts}

\usepackage{amsmath} 
\usepackage{amssymb}  
\usepackage{mathrsfs}  

\theoremstyle{plain}
\newtheorem{theorem}{Theorem}
\newtheorem{lemma}[theorem]{Lemma}
\newtheorem{definition}[theorem]{Definition}

\newtheorem{fact}[theorem]{Fact}
\newtheorem{remark}[theorem]{Remark}

\newcommand{\R}{\mathbb{R}}

\renewcommand{\d}{\mathrm{d}}

\DeclareMathOperator*{\E}{{\mathbb{E}}}

\DeclareMathOperator{\diag}{diag}

\DeclareMathOperator{\tr}{tr}

\newcommand{\mlp}{\mathrm{mlp}}
\newcommand{\attn}{\mathrm{attn}}
\newcommand{\head}{\mathrm{head}}
\newcommand{\inter}{\mathrm{inter}}

\newif\ifmodify 
\modifytrue 
\ifmodify
\newcommand{\todo}[1]{\textcolor{red}{#1}}

\else
\newcommand{\todo}[1]{}

\fi

\begin{document}

\maketitle

\input{sections/0-abstract}

\input{sections/1-introduction}

\input{sections/2-related-work}

\input{sections/3-methods}

\input{sections/4-results}

\input{sections/5-conclusion}

\bibliography{aaai25}



\clearpage
\onecolumn

\input{sections/7-appendix}
\clearpage

\input{sections/8-preliminary}

\input{sections/9-theory}

\end{document}

%% file: sections/0-abstract.tex
\begin{abstract}

Transformers have emerged as the leading architecture in deep learning, proving to be versatile and highly effective across diverse domains beyond language and image processing.
However, their impressive performance often incurs high computational costs due to their substantial model size. 
This paper focuses on compressing decoder-only transformer-based autoregressive models through structural weight pruning to improve the model efficiency while preserving performance for both language and image generation tasks.
Specifically, we propose a training-free pruning method that calculates a numerical score with Newton's method for the Attention and MLP modules, respectively.
Besides, we further propose another compensation algorithm to recover the pruned model for better performance.
To verify the effectiveness of our method, we provide both theoretical support and extensive experiments.
Our experiments show that our method achieves state-of-the-art performance with reduced memory usage and faster generation speeds on GPUs.

\end{abstract}

%% file: sections/1-introduction.tex
\section{Introduction}

Transformers have been dominant in generative models.
This includes Large Language Models (LLMs)~\cite{vaswani2017attention, touvron2023llama2} for language generation, as well as recent autoregressive image generation models~\cite{van2017vqvae, esser2021vqgan, ramesh2021dalle,yu2022sparti}.
Notably, models such as LlamaGen~\cite{sun2024llamagen}, which use image tokenizers to convert continuous images into discrete tokens, have demonstrated the ability to surpass diffusion models~\cite{ho2020denoising, rombach2022stablediffusion} in image generation tasks.
The ``\textit{next-token prediction}'' paradigm demonstrates significant capabilities in addressing both language and image generation tasks, enabling solutions that mimic human-like conversational interactions~\cite{ achiam2023gpt, li2024autoregressive}.

Recognizing the capabilities of large autoregressive models 
pioneering works~\cite{frantar2023sparsegpt, sun2023wanda, ma2023llmpruner, ashkboos2024slicegpt,zhan2021achieving,zhao-etal-2024-pruning,zhan2024fast} have sought to compress these models to enhance their execution efficiency.
Compared to irregular pruning methods, structural pruning offers a more efficient reduction in both computational and memory overhead \cite{jian2021radio,gong2022all,gong2023condense}.
By maintaining a consistent and regular structure, it simplifies implementation, accelerates processing, and leads to more predictable resource savings \cite{kong2022spvit,kong2023peeling}.
However, most of these efforts focus solely on language models and language-related research areas.
Consequently, their methods are not readily applicable to image generation tasks because of the fundamental differences in data structure and computational requirements between language and image processing~\cite{reed2016generative, parmar2018image, lee2022autoregressive, shen2024agile, shen2024edgeqat, shen2024hotaq,shen2023deepmad,shen2022data,yize2023less,yize2024neural,yize2024pruning}.
Therefore, it is crucial to explore the transformer architecture itself, rather than focusing on specific application models.
This motivates us to develop a general method for compressing autoregressive models applicable to multiple kinds of generative tasks.

Additionally, the recovery of pruned models are crucial.
Full-parameter retraining of large autoregressive models after pruning is often computationally prohibitive, making calibrations with a few samples a preferred approach.
Previous work~\cite{frantar2023sparsegpt} employs the Optimal Brain Surgeon (OBS) technique~\cite{obs1, obs2} for weight updates during pruning.
However, its heavy reliance on the approximation
information increases sensitivity to noise and reduces robustness across different datasets.
SliceGPT~\cite{ashkboos2024slicegpt} relies on a large number of samples for pruning and calibration, leading to overfitting on  calibration data and limiting the generalization 
to other different datasets.


In this work, we present a novel structural pruning approach that leverages our proposed numerical score, combined with compensation techniques for performance recovery.
We first calculate the numerical score for each layer through solving the optimal pruning mask for the minimization of pruning errors using the Newton's method.
By ranking these numerical scores of all layers, we generate the globally pruning mask with the specified pruning ratio. 
Additionally, we introduce a compensation algorithm to recover pruned models by updating the remaining weights to account for the loss caused by the pruned weights.
We empirically evaluate our method using the LLaMA model family including LLaMA, LLaMA-2, and LLaMA-3 as representative LLMs and LlamaGen for image generation tasks.
Experimental results show that our method outperforms other state-of-the-art approaches in both language and image generation tasks, validating the effectiveness of our proposed numerical score and compensation algorithm.
Moreover, 
our method reduces GPU memory usage and accelerates generation without requiring any additional GPU-specific modifications.
Our main contributions  are summarized as follows,
\begin{itemize}
    \item We propose a numerical score, derived from the numerical solution of the optimal mask for minimizing pruning errors with Newton's method.
    
    \item We propose a compensation algorithm for the reconstruction of the pruned model, further enhancing the task performance of the pruned model.

    \item Experimental results show that our method not only achieves state-of-the-art performance but also reduces memory usage and accelerates generation on GPUs.
    
\end{itemize}

%% file: sections/2-related-work.tex
\section{Related Work}

\subsection{Compression for LLMs}
The large number of parameters in LLMs motivates the need for pruning \cite{gong2020privacy,wu2022compiler,zhan2024exploring,li2022pruning,zhang2022advancing,zhan-etal-2024-rethinking-token,shen2024search} to improve efficiency. 
The work~\cite{frantar2023sparsegpt} introduces the Optimal Brain Surgeon (OBS) method~\cite{obs1, obs2} to compress the LLMs, which removes weights with minimal impact on the loss function.
It then updates the remaining weights by utilizing the inverse of the Hessian matrix to mitigate errors caused by the pruning process.
Unfortunately, this kind of pruning method is still irregular, meaning it does not lead to significant reductions in memory and computational requirements.
Subsequent works, such as LLM-Pruner~\cite{ma2023llmpruner}, SliceGPT~\cite{ashkboos2024slicegpt}, and FLAP~\cite{an2023flap}, propose structural pruning methods that effectively reduce memory usage and accelerate inference on GPUs.
These methods offer significant advantages over irregular pruning by directly enhancing the utility and efficiency of the models.
While autoregressive models excel in sequential data processing, such as text, the distinct nature of image data, where spatial relationships and pixel-level details are critical, demands different approaches.
As a result, adapting these models to image generation introduces complexities that limit their scalability and effectiveness.


\subsection{Autoregressive Models in Image Generation}
Autoregressive models, initially renowned for their success with LLMs, have recently gained popularity in the image generation research area.
Pioneering works~\cite{van2017vqvae, esser2021vqgan} introduced image tokenizers that convert continuous images into discrete tokens.
These tokenizers, which have been demonstrated to be effective by the following works~\cite{ramesh2021dalle, yu2021vector, yu2022sparti}, enable autoregressive models to generate image tokens using the next-token prediction approach.
Recent work~\cite{sun2024llamagen} delivers a series of image generation models with a new constructed image tokenizer.
This research demonstrates the effectiveness of LLM frameworks in image generation tasks, validating their potential beyond traditional language applications.
Additionally, the work~\cite{li2024autoregressive} delves deeper into the continuous-valued domains of autoregressive models and removes the image tokenizers for image generation tasks.
This work achieves stronger results while leveraging the speed advantage of sequence modeling, which further enhances the utilization and demonstrates the potential of autoregressive models in image generation tasks.


%% file: sections/3-methods.tex
\begin{figure}[t]
  \centering
  \includegraphics[width=1.0\linewidth]{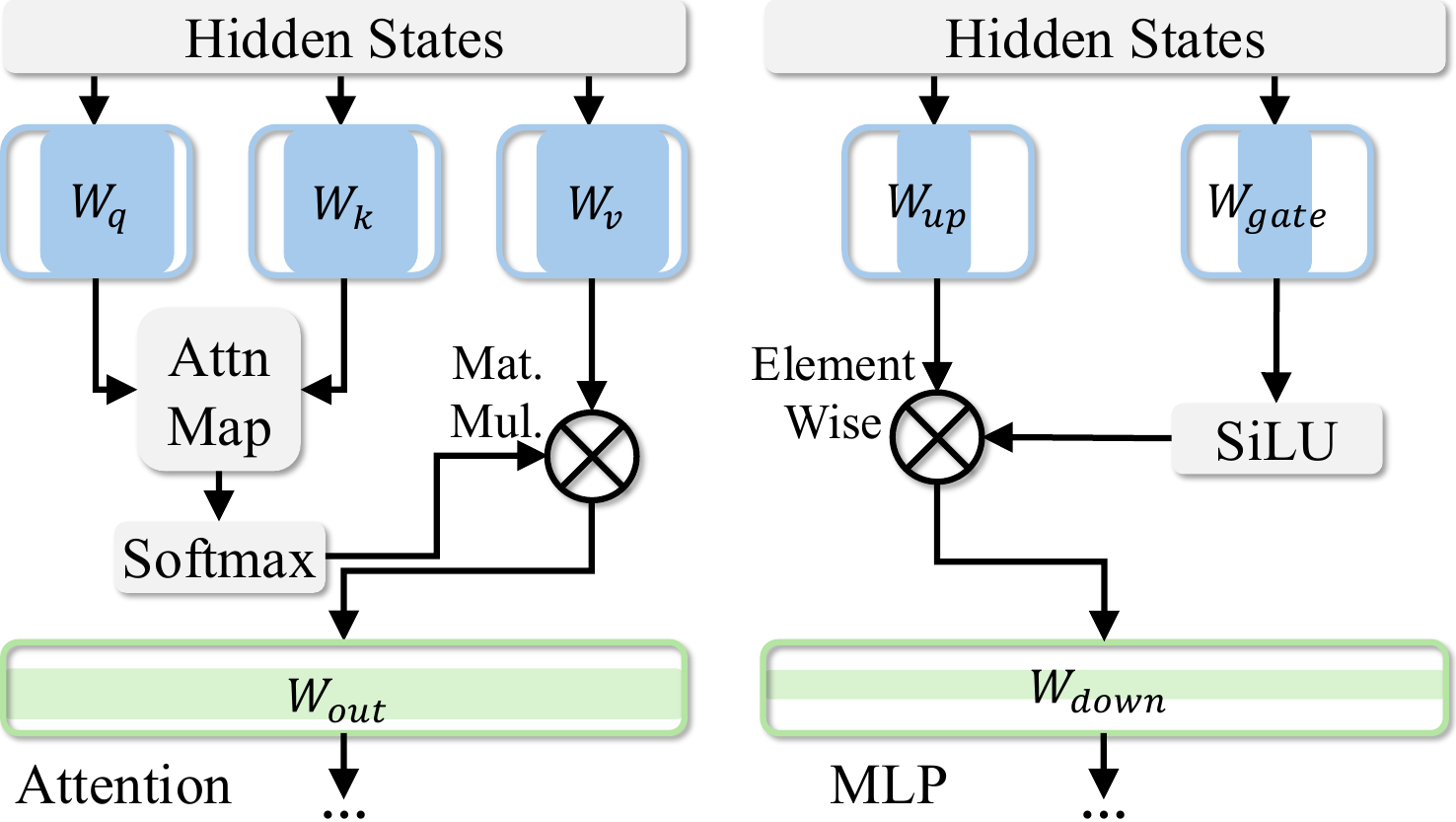}
  \caption{
  Pruning overview.
  Blue modules denote column pruning and green modules denote row pruning.
  }
  \label{fig:prune_overview}
\end{figure}

\section{Methodology}

\subsection{Preliminary}

\paragraph{Notations.}
We use $\E[\cdot]$ to denote the expectation. 
For two vectors $x \in \mathbb{R}^n$ and $y \in \mathbb{R}^n$, we use $\langle x, y \rangle$ to denote the inner product between $x,y$, \ie, $\langle x, y\rangle = \sum_{i=1}^n x_i y_i$.
We use ${\bf 1}_n$ to denote a length-$n$ vector where all the entries are ones.
We use $x_{i,j}$ to denote the $j$-th coordinate of $x_i \in \mathbb{R}^n$.
We use $\|x\|_p$ to denote the $\ell_p$ norm of a vector $x \in \mathbb{R}^n$. 
For each $a, b \in \mathbb{R}^n$, we use $a \circ b \in \mathbb{R}^n$ to denote the vector where $i$-th entry is $(a\circ b)_i = a_i b_i$ for all $i \in [n]$.
We use $\|A\|$ to denote the spectral norm for matrix $A$. 
For a square matrix $A$, we use $\tr[A]$ to denote the trace of $A$, \ie, $\tr[A] = \sum_{i=1}^n A_{i,i}$. 

\paragraph{Internal Computation Alignment.}
To maintain model interpretability and mitigate the risk of model drift, we ensure consistency in the internal computations of the Attention module. 
This approach is aligned with established methodologies in the literature~\cite{vaswani2017attention, NEURIPS2020_61d77652}.
Considering the definition in this paper mainly focuses on $X\cdot W$ where $X \in \mathbb{R}^{N\times D}$ and $W \in \mathbb{R}^{D \times D'}$, we visualize the pruning strategy in Figure~\ref{fig:prune_overview}.
In detail, we utilize the identical pruning mask (\ie, pruning strategy) for the columns of weights associated with the query, key, and value, as well as for the rows of weights in the output projection. 
Meanwhile, we apply the same strategy to the MLP module, using column pruning for the up and gate weights, and row pruning for the down projection weights.
In this paper, we construct structural pruning metrics focusing on the output projection layers of Attention module and the down projection layers of MLP module.

    

\subsection{Numerical Score}\label{sec:numerical_score}
We define the weight as $W \in \R^{D \times D'}$, input as $X \in \R^{N \times D}$, and we denote the mask as $M \in \{0,1\}^{D}$.
Additionally, we define the pruning ratio $\rho \in [0,1]$ as the ratio of the number of zeros to the total number of entries in pruning mask $M$. 

Note that, when we apply the mask column by column, the mask $M$ is a $D$-dimensional vector. Specifically, if $M_j = 0$ for $j \in [D]$, we prune the entire row for $W$, \ie, $W_{j} = 0$, and if $M_j = 1$ we keep the original $W_j$. 





To compute the numerical score, we explore the bound of the error (\ie, difference) between the original weights and pruned weights. For the bound of the error, we first formulate the error for $i \in [D']$ as 
\begin{align}\label{eq:error}
    \|X W_{*,i} - X(M \circ W_{*,i})\|_2.
\end{align}
Simplify further, for $i \in [D']$, Eq.~\eqref{eq:error} can be transformed into the following,
\begin{align*}
    \|X (({\bf 1}_D - M) \circ W_{*,i})\|_2.
\end{align*}
In the above equation, the $\|{\bf 1}_D - M\|_2$ denotes the number of zero entries in $M$, which is corresponding to the simply $\rho \cdot D $. 
Furthermore, assuming $\|X\| \leq R$, we demonstrate that the following Lemma~\ref{lem:XW_upper_bound:informal} holds,
\begin{lemma}[informal version of Lemma 9 at Appendix D.2]\label{lem:XW_upper_bound:informal}
We show that for $i \in [D']$ we have
\begin{align*}
    \|X W_{*,i} - X(M \circ W_{*,i})\|_2 \leq \rho R \| W_{*,i}\|_2.
\end{align*}
\end{lemma}
It is intuitive for us to minimize the error after establishing the error bound in Lemma~\ref{lem:XW_upper_bound:informal}.
Thus, we examine each term.
In the error bound of Lemma~\ref{lem:XW_upper_bound:informal}, the pruning ratio $\rho$ is manually specified. 
We adopt the normalization for the input $X$, then the norm of normalized $X$ is upper bounded by $1$.
Meanwhile, for $\|W_{*,i}\|_2$ term, it is the $\ell_2$ norm for $i$-th column of weight $W$.

In order to minimize the error, we regulate both $\rho$ and $\|W_{*,i}\|$.
Then, we generalize the mask $M$ from binary value to real value for the calculation of the numerical score.
Meanwhile, we set one threshold which converts the real-valued mask back into a binary mask.
For mask $M \in [0,1]^{D}$ and pruning ratio $\rho \in [0,1]$, the calculation of the numerical score is formulated as follows,
\begin{align}\label{eq:get_mask}
    \underset{M}{\arg \min} & ~~~~ \sum_{i \in [D']} \|X W_{*,i} - X(M \circ W_{*,i})\|_2, \\
    \text{s.t.} & ~~~~ \langle {\bf 1}_D, M \rangle = (1-\rho)D \notag.
\end{align}
To better solve Eq.~\eqref{eq:get_mask}, we define the numerical score $z \in [0,1]^D$ and $r := (1-\rho)D \in [0,D]$.
The equality constraint in Eq.~\eqref{eq:get_mask} is then equivalent to $\langle {\bf 1}_D , z \rangle - r = 0$.

\begin{algorithm}[!ht]\caption{Numerical Score with Newton's Method}\label{alg:main}
\begin{algorithmic}[1]
\Procedure{NumericalScore}{$X \in \R^{N \times D}, W \in \R^{D \times D'}, r \in [0,D], \lambda \in \R_+, T \in \mathbb{N}_+$}
\Comment{Theorem~\ref{thm:mask_optimize:informal}} 
    \State We choose the initial point $z_0$ such that $z_0 \in [0,1]^D$
    \For{$t=0 \to T$} 
        \State $g_l \gets ((W W^\top) \circ (X^\top X)) (z - {\bf 1}_D)$
        \State $g_r \gets \lambda (\langle {\bf 1}_D , z \rangle - r )  \cdot {\bf 1}_D$
        \State $g \gets g_l + g_r \in \R^D$ 
        \State $H_l \gets (W W^\top) \circ (X^\top X)$
        \State $H_r \gets \lambda  \cdot {\bf 1}_{D \times D}$
        \State $H \gets H_l + H_r \in \R^{D \times D}$ 
        \State $z_{t+1} \gets z_t - H^{-1} g$
    \EndFor
    \State $z \gets z_{T+1}$
    \State \Return $z$
\EndProcedure
\end{algorithmic}
\end{algorithm}

Then, Eq.~\eqref{eq:get_mask} becomes the minimization problem with the equality constraint. 
To efficiently solve such problem, we adopt the Newton' method~\cite{b15}.
By turning the equality constraint into a penalty term for regularization, we further generate the following equivalent problem,
\begin{align}\label{eq:mask_optim}
    \underset{z \in [0,1]^D }{\arg \min} & ~~~~ \frac{1}{2}\sum_{i \in [D']} \|X W_{*,i} - X(z \circ W_{*,i})\|_2^2 \\
    & ~~~~ + \frac{1}{2} \lambda \cdot ( \langle {\bf 1}_D , z \rangle - r )^2 \notag,
\end{align}
where $\lambda \in \R_+$ is the regularization parameter. 

To explain how we solve this, we define the loss function for $i \in [D']$ as follows,
\begin{align}\label{eq:loss}
    L(z)_i = \frac{1}{2} \|X W_{*,i} - X(z \circ W_{*,i})\|_2^2.
\end{align}
Meanwhile, for regularization term, we define as follows,
\begin{align}\label{eq:reg_loss}
    L_{\mathrm{reg}}(z) = \frac{1}{2} \lambda \cdot ( \langle {\bf 1}_D , z \rangle - r )^2.
\end{align}

Combining Lemma 12 and Lemma 13 at Appendix D.3, we compute the gradient of Eq.~\eqref{eq:loss} and Eq.~\eqref{eq:reg_loss} as follows,  
\begin{align}\label{eq:g}
    g = & ~~ \underbrace{((W W^\top) \circ (X^\top X)) (z - {\bf 1}_D)}_{\text{Gradient of }L(z)} \notag \\
    & + ~~~~ \underbrace{\lambda (\langle {\bf 1}_D , z \rangle - r )  \cdot {\bf 1}_D}_{\text{Gradient of }L_{\mathrm{reg}}(z)}   .
\end{align}

Combining Lemma 14 and Lemma 15 at Appendix D.3, we compute the Hessian of Eq.~\eqref{eq:loss} and Eq.~\eqref{eq:reg_loss} as follows,
\begin{align}\label{eq:H}
    H = ~~ \underbrace{(W W^\top) \circ (X^\top X)}_{\text{Hessian of }L(z)} + \underbrace{\lambda  \cdot {\bf 1}_{D \times D}}_{\text{Hessian of }L_{\mathrm{reg}}(z)}.
\end{align}

Subsequently, using Algorithm~\ref{alg:main}, we efficiently compute the optimal numerical $z$ in $O(TD^3)$, where $T$ represents the number of iterations for Newton's Method, typically around 50 in practice.
Besides, we derive the following Theorem~\ref{thm:mask_optimize:informal}.
\begin{theorem}[Mask optimization, informal version of Theorem 10 at Appendix D.3]\label{thm:mask_optimize:informal}
If the following conditions hold:
\begin{itemize}
    \item Let $W \in \R^{D \times D'}$, $X \in \R^{N \times D}$.
    \item Let $z \in [0,1]^{D}$.
    \item Let $r \in [0,D]$ denote the number of ones (it can be a fractional number). 
    \item Let $\lambda >0 $ denote a regularization co-efficients.
    \item Assume $\|X\| \leq R$. 
\end{itemize}
There exists an algorithm (Algorithm~\ref{alg:main}) that can get the optimal $z$ in $O(TD^3)$ for Eq.~\eqref{eq:mask_optim}.
\end{theorem}

\subsection{Global Pruning}
To ensure consistent head-level computation in the Attention module, given numerical scores $z^{\attn} \in \R^{D}$, we group the scores of channels for $h$-th head to determine the importance score $z^{\head}_h$ of individual heads as follows,
\begin{align*}
    z^{\head}_h = \frac{1}{D_h} \cdot \sum_{i =h \cdot D_h}^{(h+1) \cdot D_h}  z^{\attn}_i, 
\end{align*}
where $D_h$ denotes the dimension of each head, $h \in [H]$ is the head index.

Unlike the Attention module, where heads work in parallel to capture various aspects of the input and their outputs are interdependent, the MLP module has a simpler structure with minimal interdependencies between its components.
Thus, we retain the channel scores $z^{\mlp} \in \mathbb{R}^{D}$ to guide the pruning process for MLP module.

To ensure a balanced pruning process that reflects the relative importance of each layer, we simultaneously sort the numerical scores across all layers to derive the globally pruning mask.
Since a single head in the Attention module is evaluated with one score but contains significantly more weights than a single channel in the MLP module, we apply scaling factors based on the model design to balance number of pruned parameters between the Attention heads and MLP channels. 
For a specified global pruning ratio $\rho$, hidden state dimension $D$, head dimension $D_h$ in the Attention module, and intermediate size $D_{\inter}$ in the MLP module, we define $\Psi$ as the set that stores the scores for the whole model, then apply the scaling factor $\alpha$ when generating the threshold $\eta$ for all the scores as follows,
\begin{align}\label{eq:eta_threshold}
    \Psi := & ~ \alpha \cdot ( \{ \{ z^{\head}_{h,l} \}_{h=1}^{H} \}_{l=1}^{L} ) \bigcup \{ \{ z^{\mlp}_{i,l} \}_{i=1}^{D_{\inter}} \}_{l=1}^{L}, \\
    \eta = & ~ \mathrm{sort}(\Psi)  [(\rho (L \cdot H + L \cdot D_{\inter})) ], \ \alpha = \frac{4 D_h}{3}  , \notag
\end{align} 
where $H = \mathrm{index}(D / D_h)$ denotes the number of head in Attention module, $L$ denotes the number of layers for the whole model. Since the Attention module involves pruning 4 linear projections (query, key, value, and output) in head level, while the MLP module prunes only 3 (up, gate, and down) in channel level, the scaling factor $\alpha$ is given by $\frac{4 D_h}{3}$.

When the threshold $\eta$ is determined, we prune the heads in the Attention module and the channels in the MLP module across all layers based on the strategy that removes heads or channels with numerical scores below the threshold.

\subsection{Compensation for Pruning}\label{sec:compensation_pruning}

With the above discussion, we obtain the pruning mask with Newton's method.  To further improve the  model  performance, we  modify  the  remaining weights in the model  to compensate the loss of the pruned weights.

\paragraph{Problem Formulation.}

Note that to align the internal computations in the attention and MLP modules,  we prune  the  rows of the output layers in the modules  and the columns in other layers of the modules.   If the columns of a layer with $W$  is pruned in $ X W$,   the corresponding columns of the output  also become zero and we are not able to compensate its loss, since modifying other  unpruned columns can not change the zero output for the pruned columns.  Thus,   we only  update the weights of the output layers   with row pruning in the Attention and MLP modules. We modify the remaining rows based on pruned rows in $W$. 
For layers with column pruning, we do not modify their unpruned weights. 

For the original weights $W$, after pruning, there are $k$ pruned rows which are all zeros and their row indexes are denoted by $p_i, \forall i \in [k]$.  We  modify the weights   with  the weight perturbations $\delta  W$, so that the layer output difference (before and after pruning) measured with  $\ell_2$ norm  is minimized.  
The weight optimization problem can  formulated as the following,
\begin{align} \label{eq:new_problem}
\min_{\delta W} \ \ & ~~~~ \mathcal{L}(\delta W)  =   \|  X 
 ( W + \delta  W) - X W   \|_2^2 =  \| X \delta  W  \|_2^2 , \notag  \\ 
\text{s.t.} \ \ 
& ~~~~  e_{p_i}^\top  \delta  W + (W)_{p_i,*} =  0,  \quad \text{for } i = 1, 2, \dots, k.
\end{align}
where $ e_{p_{i}} \in  \R^{D\times 1}$  is the one-hot vector with the $p_i^{th}$ element as 1 and all others as 0.  Thus,  $  e_{p_{i}}^\top \delta W$  denotes selecting the $p_{i}^{th}$ row of $ \delta  W$. $(W)_{i,j}$ represents the element in the $i^{th}$ row and $j^{th}$ column of the matrix.  Then, $(W)_{p_{i},*}$   represents the $p_{i}^{th}$ row  of $W$.  We can see that the constraint in Eq.~\eqref{eq:new_problem}  ensures that the corresponding pruned  rows in the modified weights are all zeros, and the remaining weights are optimized to minimize the loss incurred by pruned rows.

 It can be  further transformed to  the following,
\begin{align} \label{eq:vec:problem}
\min_{\delta W} \ \ &~~~~ \mathcal{L}(\delta W)  =   \| X  \delta W   \|_2^2 ,   \notag  \\
\text{s.t.} \ \  &~~~~  M_p^\top \delta  W  + W_{p} = 0 , 
\end{align}
where  $M_p \in \R^{D\times k}$  is the collection of  all $e_{p_{i}}$, \ie,  $ (M_p)_{*, i} = e_{p_{i}}$, or  $ (M_p^\top)_{i, *} = e_{p_{i}}^\top, 
 \forall i \in [k]$.  Similarly,    $W_{p}$  is a  collection of all pruned rows in $W$ with  $(W_{p})_{ i,*} = (W)_{p_{i},*}, \forall i \in [k] $.   We have $W_{p} = M_p^\top W $. 

\begin{figure*}[t]
  \centering
  \includegraphics[width=1.0\linewidth]{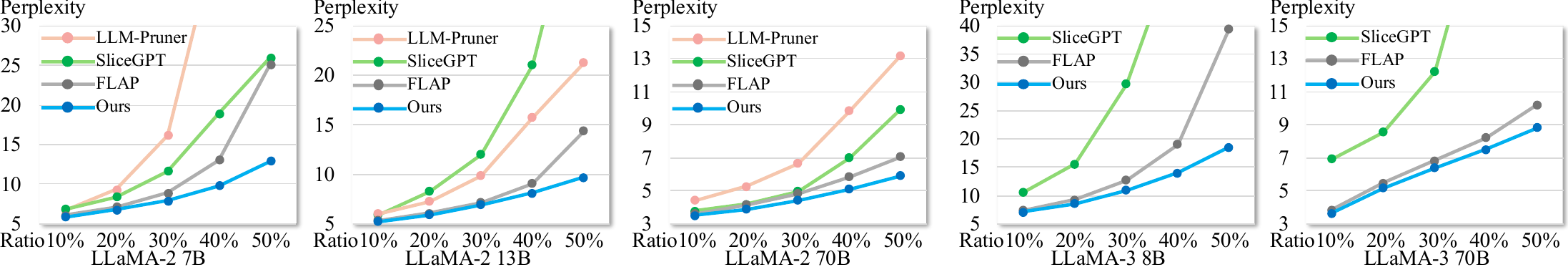}
  \caption{
  Perplexity ($\downarrow$) results for LLaMA-2 and LLaMA-3 models on WikiText2 dataset with 2048 sequence length.
  Comprehensive detailed results are included in Table 6 and Table 7 at Appendix A.2 and A.3.
  }
  \label{fig:llama2_llama3_wiki_results_visual}
\end{figure*}

\input{tables/0-main-results}

\paragraph{Optimal Solution.} 
Eq.~\eqref{eq:vec:problem} can be solved analytically with the following Theorem~\ref{thm:optimal_weight}. 
The detailed  proof is shown in  Appendix B.

\begin{theorem} \label{thm:optimal_weight}
The optimal solution for Eq.~\eqref{eq:vec:problem} can be derived as the following,
\begin{align}   \label{eq:optimal_w_final}
\delta  W^* 
= & -  (2  X^\top  X)^{-1}  M_p  (M_p^\top   (2  X^\top  X )^{-1} M_p  )^{-1}     M_p^\top   W. 
\end{align} 
\end{theorem}
\begin{remark}
The optimal loss of Problem~\eqref{eq:vec:problem} corresponding to the optimal  weight perturbation can be expressed as  
\begin{align}  \label{eq:optimal_loss}
L^* =  \frac{1}{2}  \sum_i  ( W^\top   M_p ( M_p^\top   (2  X^\top  X )^{-1}  M_p )^{-1}   M_p^\top  W )_{i,i}.
\end{align} 
The sum in Eq.~\eqref{eq:optimal_loss}  is computed over $D'$ (the number of columns in $ W$), \ie, $ i \in [D']$. 
\end{remark}
\begin{remark}
If  the rank of  $2  X^\top X  $ is not full so that  the inversion $(2  X^\top  X  )^{-1}$ is unavailable, we apply the dampening method  to compute $(2  X^\top  X  + \gamma \cdot I )^{-1} $ instead of $(2  X^\top  X  )^{-1}$, with $\gamma$ as the dampening ratio. 
\end{remark}

\subsection{Complexity Analysis}

For the computation of numerical score,  according to the Lemma~\ref{lem:XW_upper_bound:informal} and Theorem~\ref{thm:mask_optimize:informal}, the complexity is $O(TD^3)$ where $T$ represents the number of iterations for Newton's Method, typically around 50 in practice.
Additionally, for the compensation method, as demonstrated in Eq.~\eqref{eq:optimal_w_final},  the complexity is $O(D^3)$ as we need to compute the inverse of a matrix.
The matrix multiplication with $M_p$ or $M_p^T$ just selects the  columns or rows of a matrix, without the need of  actual multiplication. 
The complexity for numerical score calculation and compensation is the same with state-of-the-art methods, such as SparseGPT~\cite{frantar2023sparsegpt}. 
In practice, the compensation is finished with just a few data samples on only the output projection layers of the Attention module and the down projection layers of the MLP module, which is more efficient compared with other recovery methods such as LLM-Pruner \cite{ma2023llmpruner} to finetune the whole model on whole dataset, or SliceGPT \cite{ashkboos2024slicegpt} to adopt a large amount of samples for calibration.

%% file: tables/0-main-results.tex
\begin{table*}[!ht]
\centering
\resizebox{0.9\linewidth}{!}{
\begin{tabular}{c|c|ccc|ccc|ccc|ccc}
\toprule
\multirow{2}{*}{Method} & \multirow{2}{*}{\begin{tabular}[c]{@{}c@{}}Prune\\ Ratio\end{tabular}} & \multicolumn{3}{|c|}{LLaMA-7B} & \multicolumn{3}{|c|}{LLaMA-13B} & \multicolumn{3}{|c|}{LLaMA-30B} & \multicolumn{3}{|c}{LLaMA-65B} \\
                        &                                                                        & Wiki     & PTB     & C4      & Wiki     & PTB      & C4      & Wiki     & PTB      & C4      & Wiki     & PTB      & C4      \\ \midrule
Baseline                & \textbackslash{}                                                       & 5.68            & 27.34           & 7.08            &     5.09           &      19.23           &   6.61              &       4.10         &        16.29         &        5.98        &    3.53           &    17.61            &      5.62          \\ \midrule
LLM-Pruner              & 10\%                                                                   & 7.41            & 36.73           & 9.25            & 6.38           & 31.85           & 8.16            & 4.92           & 18.17           & 6.63           & 3.98          & 19.44          & 6.08           \\
SliceGPT                & 10\%                                                                   & 6.97            & 88.48           & 23.54           & 6.11           & 60.15           & 20.18           & 5.24           & 39.72           & 17.83          & 4.57          & 36.20          & 14.14          \\
FLAP                    & 10\%                                                                   & 6.34            & 32.39           & 8.058           & 5.45           & 20.99           & 7.33            & 4.52           & 17.29           & 6.49           & 3.91          & 19.35          & 6.04           \\
Ours                    & 10\%                                                                   & \textbf{6.01}   & \textbf{31.65}  & \textbf{7.94}   & \textbf{5.38}  & \textbf{20.52}  & \textbf{7.27}   & \textbf{4.43}  & \textbf{17.26}  & \textbf{6.47}  & \textbf{3.82} & \textbf{19.28} & \textbf{6.02}  \\ \midrule
LLM-Pruner              & 20\%                                                                   & 10.73           & 59.73           & 12.15           & 6.38           & 31.85           & 9.42            & 5.83           & 20.18           & 7.55           & 4.65          & 21.85          & 6.75           \\
SliceGPT                & 20\%                                                                   & 8.42            & 120.89          & 35.93           & 7.17           & 86.26           & 29.70           & 6.18           & 50.95           & 26.85          & 5.34          & 61.09          & 21.86          \\
FLAP                    & 20\%                                                                   & 7.40            & 36.77           & 9.99            & 6.03           & 23.33           & 8.42            & 5.18           & 19.30           & 7.42           & 4.45          & 21.45          & 6.75           \\
Ours                    & 20\%                                                                   & \textbf{6.60}   & \textbf{35.75}  & \textbf{9.49}   & \textbf{5.89}  & \textbf{23.11}  & \textbf{8.39}   & \textbf{4.92}  & \textbf{18.58}  & \textbf{7.36}  & \textbf{4.26} & \textbf{20.94} & \textbf{6.73}  \\ \midrule
LLM-Pruner              & 30\%                                                                   & 18.58           & 93.24           & 17.78           & 11.81          & 45.42           & 12.65           & 7.59           & 24.97           & 9.08           & 5.52          & 26.38          & 7.53           \\
SliceGPT                & 30\%                                                                   & 12.75           & 258.90          & 67.33           & 9.18           & 125.40          & 46.46           & 7.74           & 75.89           & 42.71          & 6.56          & 74.43          & 35.68          \\
FLAP                    & 30\%                                                                   & 9.18            & 47.35           & 13.08           & 6.97           & 27.36           & 10.01           & 6.28           & 21.88           & 8.53           & 5.10          & 23.91          & 7.59           \\
Ours                    & 30\%                                                                   & \textbf{7.56}   & \textbf{41.05}  & \textbf{11.53}  & \textbf{6.57}  & \textbf{26.27}  & \textbf{9.98}   & \textbf{5.46}  & \textbf{20.48}  & \textbf{8.46}  & \textbf{4.75} & \textbf{22.13} & \textbf{7.51}  \\ \midrule
LLM-Pruner              & 50\%                                                                   & 126.0          & 460.7          & 73.88           & 45.69          & 152.99          & 36.94           & 19.68          & 78.29           & 18.64          & 9.34          & 43.79          & 12.16          \\
SliceGPT                & 50\%                                                                   & 1540         & 6364         & 4847         & 18.75          & 277.34          & 122.5          & 15.60          & 195.4          & 118.5         & 12.01         & 160.3         & 92.66          \\
FLAP                    & 50\%                                                                   & 21.89           & 135.8          & 30.86           & 12.88          & 53.54           & 18.37           & 13.41          & 47.30           & 13.17          & 6.98          & 28.52          & 10.36          \\
Ours                    & 50\%                                                                   & \textbf{11.66}  & \textbf{82.55}  & \textbf{20.72}  & \textbf{8.91}  & \textbf{37.56}  & \textbf{16.12}  & \textbf{7.25}  & \textbf{26.68}  & \textbf{11.91} & \textbf{6.02} & \textbf{25.17} & \textbf{9.73}  \\ 
\midrule
LLM-Pruner              & 70\%                                                                   & 9010         & 4111         & 2655         & 5900        & 6039         & 1334         & 895.7         & 3274         & 456.9         & Nan           & Nan            & Nan            \\
SliceGPT                & 70\%                                                                   & 3605         & 7304         & 8096         & 67.65          & 874.9          & 537.4          & 71.25          & 633.1          & 406.6         & 102.4        & 863.9         & 662.8         \\
FLAP                    & 70\%                                                                   & 577.9          & 1835         & 833.7          & 647.8         & 1588         & 975.1          & 2786        & 2735         & 2416        & Nan           & 2333        & Nan            \\
Ours                    & 70\%                                                                   & \textbf{162.9} & \textbf{721.3} & \textbf{361.6} & \textbf{41.66} & \textbf{275.7} & \textbf{115.3} & \textbf{39.88} & \textbf{124.2} & \textbf{50.43} & \textbf{9.65} & \textbf{69.49} & \textbf{20.84} \\
\bottomrule
\end{tabular}
}
\caption{
Perplexity ($\downarrow$) results for LLaMA-1 family models with different pruning ratios on WikiText2, PTB, and C4 with 2048 sequence length.
Full results with larger sparsity ratios are included in Table 5 at Appendix A.1.
}
\label{tab:main_llama_1_results}
\end{table*}

%% file: sections/4-results.tex
\input{tables/1-zeroshot-results}

\begin{figure*}[t]
  \centering
  \includegraphics[width=1.0\linewidth]{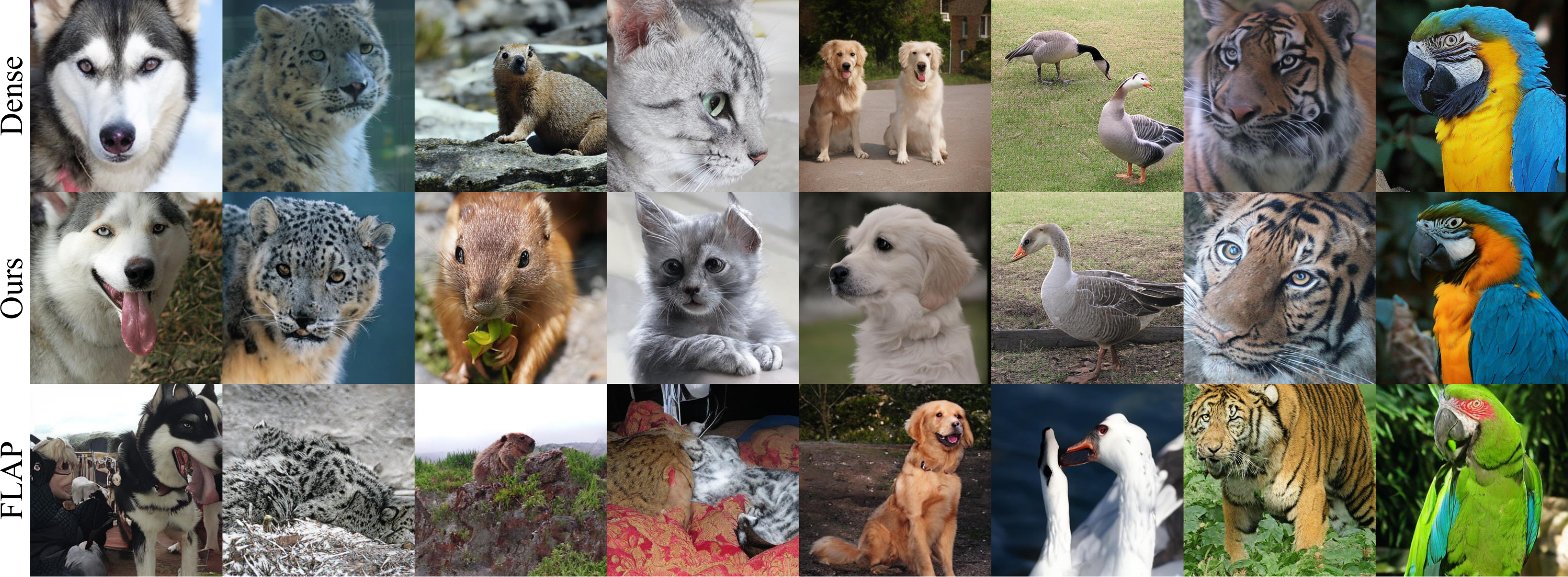}
  \caption{
  Visualization of generated images through LlamaGen-3B in 384$\times$384 resolution (cfg=1.65) with 10\% sparsity.
  }
  \label{fig:llamagen_results_visual}
\end{figure*}

\section{Experimental Results}

\subsection{Experiment Setup}

We conduct the experiments on LLaMA model families including LLaMA-1~\cite{touvron2023llama1}, LLaMA-2~\cite{touvron2023llama2}, and LLaMA-3~\cite{meta2023llama3} for the language generation tasks.
For evaluations, we compare the perplexity of the models on the WikiText2~\cite{wikitextdataset}, PTB~\cite{ptbdataset}, and C4~\cite{c4dataset} datasets with the 2048 sequence length.
We also follow LLM-Pruner to evaluate the zero-shot accuracy on common sense reasoning zero-shot classification datasets including BoolQ~\cite{clark2019boolq}, PIQA~\cite{bisk2020piqa}, HellaSwag~\cite{zellers2019hellaswag}, WinoGrande~\cite{sakaguchi2021winogrande}, ARC-easy~\cite{clark2018arc}, ARC-challenge~\cite{clark2018arc}, and OpenbookQA~\cite{mihaylov-etal-2018-suit-opqa}.
For experiments, we adopt 128 samples from training dataset of WikiText2 to compute the numerical score and compensate the pruned models.
For fairness, we also adopt 128 samples for other methods.

As for the image generation tasks, we adopt the LlamaGen~\cite{sun2024llamagen} model family with LlamaGen-XXL and LlamaGen-3B to verify the effectiveness of our method on image generation tasks.
We adopt the Fréchet inception distance (FID)~\cite{heusel2017fid}, Inception Score (IS)~\cite{salimans2016is}, sFID~\cite{nash2021sfid}, and Precision/Recall~\cite{Kynkaanniemi2019precisionrecall} as the evaluation metrics on ImageNet dataset~\cite{imagenet}.
For all evaluations, we utilized ADM’s TensorFlow scripts~\cite{dhariwal2021evalpipelie} to ensure fair and consistent comparisons.
Given that LLM-Pruner requires a backward process and SliceGPT has slow pruning, we further implement FLAP for comparative analysis in image generation tasks.
In pratical, we generate 128 images for each class of ImageNet with LlamaGen models for the computation of numerical score and compensation.
Same strategy for FLAP for fairness.

\subsection{Results of LLMs}

For the LLaMA models, we present the results with different pruning ratios varying from 10\% to 70\% in Table~\ref{tab:main_llama_1_results}. 
Based on the perplexity results evaluated with 2048 sequence length on three datasets, our method consistently outperforms other methods across all pruning ratios, demonstrating the effectiveness of our proposed approach.
Full results with more sparse ratios are included in Table 5 of Appendix A.1.
Results show that for the larger model LLaMA-65B with pruning ratio of 70\%, both LLM-Pruner and FLAP fail to produce an effective pruned model with their respective methods. 
In contrast, our method successfully maintains the most of the model's capabilities.

We further evaluate the zero-shot capabilities of the pruned model across seven downstream tasks. The results of LLaMA-7B model are shown in Table~\ref{tab:zeroshot_results}.
Full results, including additional pruning ratios and the LLaMA-13B model, are detailed in Table 9 in  Appendix A.5.
Our method demonstrates superior performance compared to the other three methods on those common sense reasoning zero-shot classification datasets.
Besides, we show the results with LLaMA and LLaMA-2 models of our method on MMLU~\cite{mmludataset} and GSM8K~\cite{cobbe2021gsm8kdataset} datasets in 
Table 8 of Appendix A.4,
which demonstrates that our method retains both generative and mathematical capabilities.

We show the results for LLaMA-2 and LLaMA-3 models with 2048 sequence length on WikiText2 dataset in Figure~\ref{fig:llama2_llama3_wiki_results_visual}.
The detailed perplexity results for both model families on three datasets are shown in Table 6 and Table 7 of Appendix A.2 and A.3.
The blue line representing our method's results consistently appears at the lowest position on the graphs, indicating its superior performance compared to the other methods with all model families.


\subsection{Results of Image Generation}

We implement the FLAP pruning method on LlamaGen model and compare this method on image generation task.
We show the sparse results with LlamaGen-XXL (1.4B) and LlamaGen-3B models on ImageNet with 384$\times$384 resolution in Table~\ref{tab:llamagen_results}.
We observe that for the smaller model LlamaGen-XXL (1.4B), our method shows a distinct advantage at higher pruning ratios.
For the larger model LlamaGen-3B, our method consistently outperforms across all pruning ratios, effectively preserving most of the original model's capabilities.
We further visualize the images generated by 10\% sparsity models in Figure~\ref{fig:llamagen_results_visual} with additional visualizations provided in 
Figure 6 of Appendix A.6.
We observe that our method generates better image results compared to FLAP method in most cases.

\subsection{Ablation Study}

\paragraph{Results with 128 Sequence Length.}
To demonstrate the effectiveness of our method for short sequence lengths, we present the results generated with a sequence length of 128 in Table~\ref{tab:128_sequence_length} using the LLaMA-7B model and the WikiText2 dataset.
Comprehensive results, including additional pruning ratios and datasets, are provided in Table 10 of Appendix A.7.
As observed, our method consistently performs the best across all pruning ratios. 

\begin{figure}[t]
  \centering
  \includegraphics[width=0.92\linewidth]{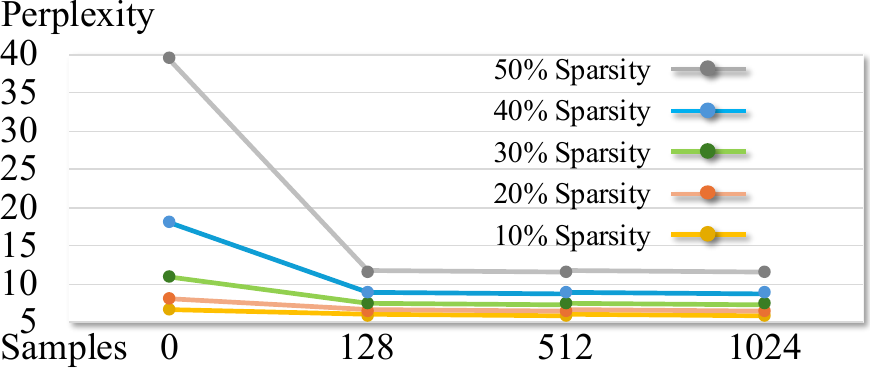}
  \caption{
  Ablation for number of samples for compensation.
  }
  \label{fig:calibration_sample}
\end{figure}

\paragraph{Number of Samples for Compensation.}
To verify the efficiency of the compensation process for our method, we conducted experiments using different numbers of samples. 
The results of these experiments are shown in Figure~\ref{fig:calibration_sample}.
The results demonstrate that the performance difference between compensation with 128 samples versus 512 or even 1024 samples is minimal across all pruning ratios.
This indicates that 128 samples are sufficient for our compensation method, highlighting its efficiency.

\input{tables/5-llamagen-results}

\input{tables/2-128-length-main-results}

\begin{figure}[!ht]
  \centering
  \includegraphics[width=0.92\linewidth]{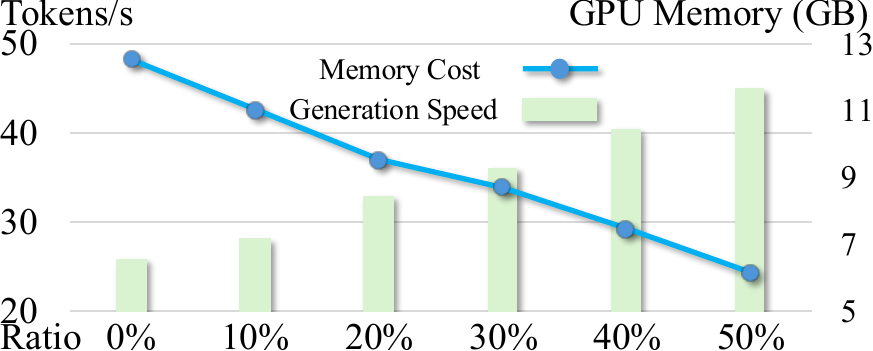}
  \caption{
  GPU memory v.s. generation speed.
  }
  \label{fig:memory_speed}
\end{figure}

\paragraph{Memory \& Generation Speed.}
We show the memory reduction and generation acceleration in Figure~\ref{fig:memory_speed}.
The results are obtained using an NVIDIA A100 GPU with a sentence consisting of 64 tokens as the model input.
The results show that as the pruning ratio increases, there is a corresponding decrease in GPU memory usage and an increase in generation speed, which validates the effectiveness of our method.

%% file: tables/1-zeroshot-results.tex
\begin{table*}[!ht]
\centering
\resizebox{0.9\linewidth}{!}{
\begin{tabular}{c|c|ccccccc|c}

\toprule
Method & Prune Ratio   & BoolQ & PIQA  & Hella Swag & Wino Grande & ARC-e  & ARC-c  & OBQA  & Average Acc.         \\
\toprule
LLaMA-7B  & /          & 73.18 & 78.35 & 72.99           & 67.01      & 67.45          & 41.38       & 42.40 & 63.25           \\
\toprule
LLM-Pruner(v)         & \multirow{3}{*}{20\%}         & 61.44 & 71.71 & 57.27           & 54.22      & 55.77          & 33.96       & 38.40 & 53.25           \\
LLM-Pruner(e2)        &         & 59.39 & 75.57 & 65.34           & 61.33      & 59.18          & 37.12       & 39.80 & 56.82           \\
LLM-Pruner(e1)       &           & 57.06 & 75.68 & 66.80            & 59.83      & 60.94          & 36.52       & 40.00   & 56.69           \\
                            \midrule
SliceGPT      & 20\%          & 37.89 & 64.09 & 45.67           & 62.75      & 53.62          & 31.74       & 33.20 & 46.99          \\
                            \midrule
FLAP           & 20\%            & 68.59 & 74.21 & 64.98           & 64.40       & 59.89          & 37.80        & 40.20 & 58.58           \\
\midrule
Ours            & 20\%     & 67.92 & 74.76 & 67.31 & 66.54 & 58.80 & 36.77 & 39.4  & \textbf{58.79}  \\
\bottomrule

\end{tabular}
}
\caption{
Pruning results for LLaMA-7B on common sense reasoning datasets.
LLM-Pruner (v) and (e$i$) denote vector-wise and element-wise with $i$-th order ($i=1,2$).
Full results with more sparsity ratios and LLaMA-13B are in Table 9 at Appendix B.
}
\label{tab:zeroshot_results}
\end{table*}

%% file: tables/5-llamagen-results.tex
\begin{table}[t]
\centering
\resizebox{1.0\linewidth}{!}{
\begin{tabular}{c|c|ccccc}
\toprule
Method   & Ratio    & FID $\downarrow$    & sFID $\downarrow$             & IS $\uparrow$     & Prec $\uparrow$ & Rec $\uparrow$  \\
\midrule
\multicolumn{7}{c}{LlamaGen-XXL (cfg=1.75)}                                                       \\
\midrule
/ & /                & 2.39   & 6.02 & 253.16  & 80.73\%      & 59.60\%    \\
\midrule
FLAP     & 10\%             & 7.87           & 9.92             & 145.25          & 61.96\%          & 63.34\%          \\
Ours     & 10\%             & \textbf{6.09}  & \textbf{7.70}    & \textbf{168.96} & \textbf{70.98\%} & \textbf{65.01\%} \\
\midrule
FLAP     & 15\%             & 15.93          & 11.81            & 100.48          & 52.05\%          & 62.02\%          \\
Ours     & 15\%             & \textbf{11.29} & \textbf{9.85}    & \textbf{124.31} & \textbf{62.95\%} & \textbf{65.84\%} \\
\midrule
FLAP     & 20\%             & 53.86          & 20.63            & 32.41           & 28.31\%          & 67.14\%          \\
Ours     & 20\%             & \textbf{22.45} & \textbf{14.16}   & \textbf{78.64}  & \textbf{53.68\%} & \textbf{67.66\%} \\
\midrule
\multicolumn{7}{c}{LlamaGen-3B (cfg=1.65)}                                                                                         \\
\midrule
/ & / & 2.26           & 6.19 & 260.46          & 82.07\%             & 58.35\%             \\
\midrule
FLAP     & 10\%             & 7.57           & 8.40             & 158.74          & 67.19\%          & 64.90\%          \\
Ours     & 10\%             & \textbf{3.97}  & \textbf{7.45}    & \textbf{202.93} & \textbf{73.93\%} & 62.86\% \\
\midrule
FLAP     & 15\%             & 38.45          & 23.61            & 57.29           & 43.45\%          & 63.27\%          \\
Ours     & 15\%             & \textbf{8.92}  & \textbf{10.32}   & \textbf{152.36} & \textbf{66.83\%} & \textbf{63.97\%} \\
\midrule
FLAP     & 20\%             & 162.15         & 93.98            & 5.97            & 13.36\%          & 28.03\%          \\
Ours     & 20\%             & \textbf{20.16} & \textbf{16.29}   & \textbf{95.05}  & \textbf{54.87\%} & \textbf{63.72\%} \\
\bottomrule
\end{tabular}
}
\caption{
Sparse results for image generation task with LlamaGen model family in 384$\times$384 resolution.
}
\label{tab:llamagen_results}
\end{table}

%% file: tables/2-128-length-main-results.tex

\begin{table}[!ht]
\centering
\resizebox{0.89\linewidth}{!}{
\begin{tabular}{cccccc}
\toprule
Prune Ratio & 10\% & 20\% & 30\% & 40\% & 50\%  \\
\midrule
LLM-Pruner  & 15.37    & 19.09    & 30.64   & 52.28   & 122.8 \\
SliceGPT    & 14.52    & 19.27    & 44.96   & 535.5   & 2241  \\
FLAP        & 13.84    & 14.62    & 17.62   & 22.32   & 31.80 \\
Ours        & \textbf{13.31} & \textbf{14.47} & \textbf{16.40} & \textbf{19.04} & \textbf{23.32}  \\
\bottomrule
\end{tabular}
}
\caption{
Results for LLaMA-7B model on WikiText2 with 128 sequence length.
Full results with more datasets are in Table 10 at Appendix B.
}
\label{tab:128_sequence_length}
\end{table}

%% file: sections/5-conclusion.tex
\section{Conclusion and Limitation}

In this paper, we propose the numerical score which is calculated through Newton's Method for the minimization of pruning errors.
We further sort numerical scores across all model layers for global pruning.
Additionally, we introduce a compensation algorithm to reconstruct weights in pruned models.
Experimental results show that our method achieves the state-of-the-art performance, which demonstrates the effectiveness of our method.
Meanwhile, our method reduces memory usage and accelerates generation on GPUs without requiring additional implementations.
One limitation of our method is its reduced effectiveness with smaller LlamaGen models for image generation tasks, primarily due to the usage of the discrete image tokenizer, which tends to lose important details as the sparsity increases.

%% file: sections/7-appendix.tex
\clearpage

\appendix

\begin{center}
    \noindent\textbf{\huge Appendix}
\end{center}

\section{More Experimental Results}
\subsection{Full LLaMA Results}\label{app:sec:llama_full}

We show the full results of LLaMA family models on three different datasets in Table~\ref{tab:main_llama_1_results_appendix} varying sparsity ratio from 10\% to 70\% with all kinds of model size.
Our method consistently achieves better performance than all other three state-of-the-art methods.

\input{tables/0.2-llama-1-results-appendix}

\clearpage

\subsection{LLaMA-2 Results}   \label{app:sec:llama2_full}

We show the detailed perplexity results of LLaMA-2 family models on three different datasets in Table~\ref{tab:llama_2_results}.
Our methods achieves better performance and shows better generalization on different datasets than all the other methods.

\input{tables/3-llama_2-results}

\clearpage

\subsection{LLaMA-3 Results}   \label{app:sec:llama3_full}

We further show the detailed perplexity results of LLaMA-3 family models on three different datasets in Table~\ref{tab:llama_3_results}.
LLM-Pruner codebase does not support LLaMA-3 models.
We achieve consistent better performance than all the other models on three datasets with all kinds of model sizes.

\input{tables/4-llama_3-results}

\subsection{Results on Generation and Math Datasets}  \label{app:sec:gen}

We provide the results of our method on generation and math datasets in Table~\ref{tab:mmlu_gsm8k}.
The results demonstrate that our pruning method effectively preserves both the model's generation capabilities and its mathematical performance.

\input{tables/6-mmlu-gsm8k-llama_1-llama_2}

\clearpage
\subsection{Full Results for Common Sense Reasoning Datasets}  \label{app:sec:QA}

We provide the task performance on common sense reasoning dataset with LLaMA-7B and LLaMA-13B models with 10\% and 20\% sparsity in Table~\ref{tab:zeroshot_results_full_appendix}.
The results show that our method can perform better than other three methods.

\input{tables/1-zeroshot-results-appendix}

\clearpage

\subsection{More Results of Image Generation Tasks}  \label{app:sec:gen_image}

\begin{figure*}[!ht]
  \centering
  \includegraphics[width=1.0\linewidth]{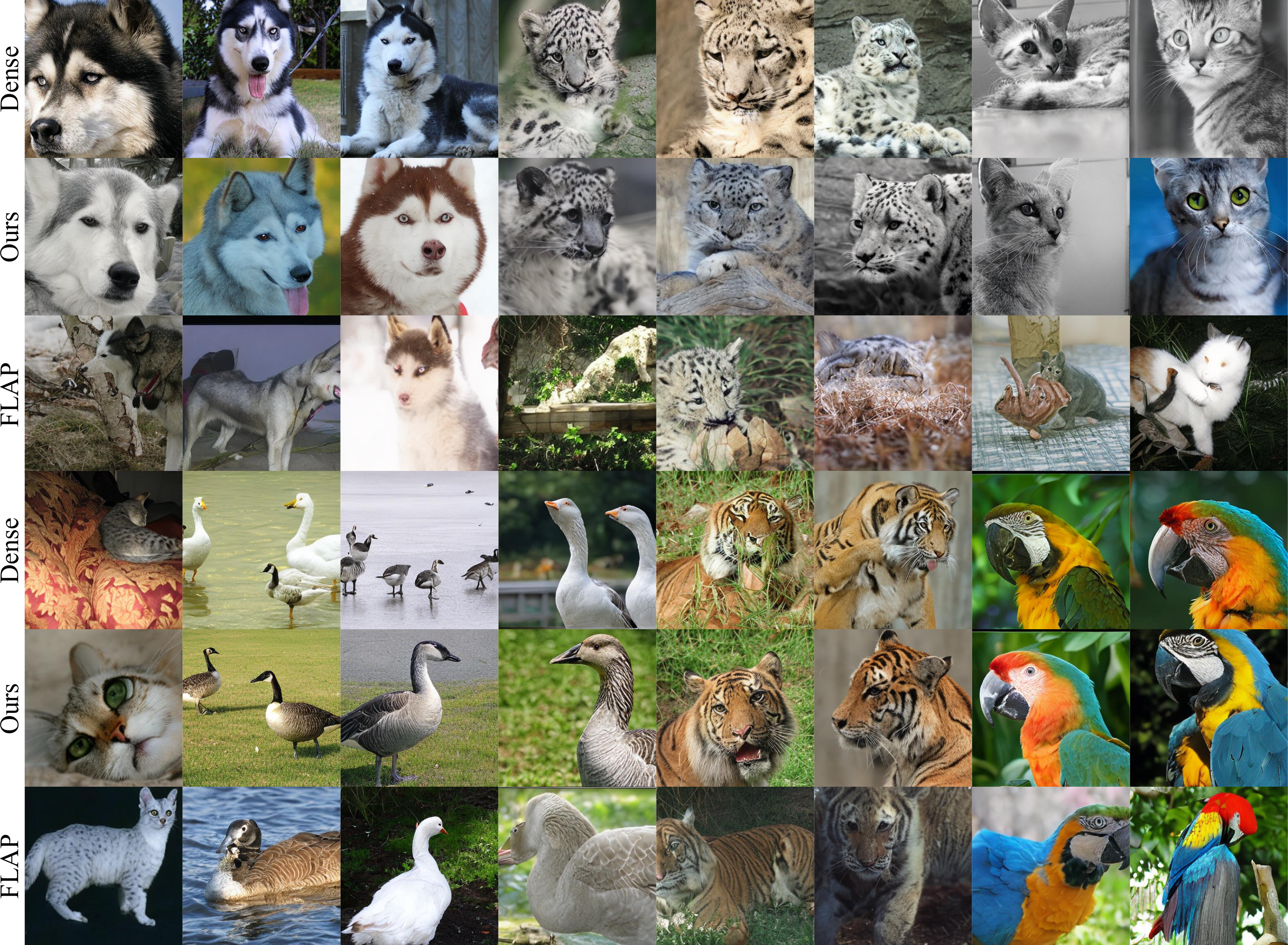}
  \caption{
  Visualization of generated images through LlamaGen-3B in 384$\times$384 resolution (cfg=1.65) with 10\% sparsity.
  }
  \label{fig:llamagen_result_visual_appendix}
\end{figure*}

We visualize more image generation results in Figure~\ref{fig:llamagen_result_visual_appendix}.



\subsection{Full Results with 128 Sequence Length}   \label{app:sec:128}

We provide the full perplexity results with 128 input sequence length on three different datasets in Table~\ref{tab:128_sequence_length_full_appendix}.
The results show the effectiveness of our method for short input sequence.

\input{tables/2-128-length-main-results-appendix}

\clearpage

\section{Derivation of Theorem 3}  \label{app:sec:proof_3}

The Lagrange function for  Problem~(\ref{eq:vec:problem}) is   
\begin{align} \label{eq:lagrange}
\mathscr{L}(\delta W,  \lambda)    =  & \|X   \delta W   \|^2   +  \sum_i    \lambda_{i}^\top  (M_p^\top  \delta  W + W_{p})  \cdot e_i \notag  \\
  =  & \tr [  \delta W^\top  X^\top X \delta W   ]   +    \sum_i    \lambda_{i}^\top  (M_p^\top  \delta W + W_{p})\cdot e_i,
\end{align}
where    $\lambda_i \in R^{k \times 1} $ denotes  the Lagrange multiplier corresponding to the constraint for the $i$-th column in Eq.~\eqref{eq:vec:problem}.  $e_i $ is a one-hot vector with the $i$-th element as 1 and all others as zero. $A  e_i$ denotes selecting the $i$-th column of a matrix. 
Note that for the constraint  in Eq.~\eqref{eq:vec:problem},  $\delta M_p^\top W + W_{p}$  is a matrix and every element in the matrix should be 0. Thus, in the Lagrange function,  we assign a Lagrange multiplier for each element in the constraint. Specifically, $\lambda_i  = [ \lambda_{i1},  \lambda_{i2}, ...,  \lambda_{ik}]^\top$  and each $\lambda_{ij}$ corresponds to the $j$-th row and $i$-th column  of  the constraint.  And their sums are computed in the Lagrange function. 
The trace function $\tr[\cdot]$ computes the $\ell_2$ norm of $X \delta W$.

The gradients with reference to $\delta W$   should be 0, \ie,
\begin{align} 
\frac{\delta \mathscr{L}(\delta W,  \lambda) }{\delta  (\delta W)} = 2   X^\top  X \delta W + \sum_i   M_p \lambda_i e_i^\top = 0.
\end{align}
 $\delta W$ can be derived  as below, 
\begin{align} \label{eq:delta_half}
\delta W   = - (2 X^\top X )^{-1} \left(  \sum_i   M_p \lambda_i  e_i^\top  \right).
\end{align}

By applying Equation (\ref{eq:delta_half})  in Eq.~\ref{eq:lagrange},  we have the following, 
\begin{align}
g(\lambda) = & ~ \tr \left[  \left(  \sum_i   e_{i} \lambda_i^\top M_p^\top \right) (2 X X^\top )^{-1} X^\top X   (2 X^\top X )^{-1} \left(  \sum_i   M_p \lambda_i   e_i^\top   \right) \right] \notag \\
& ~  - \sum_i  \lambda_i^\top  M_p^\top  (2 X^\top X )^{-1} \left(  \sum_i  M_p \lambda_i  e_i^\top   \right)  e_i   + \sum_i  \lambda_i^\top  W_{p}  e_i 
 \notag \\
= & ~   \tr \left[X(2 X^\top X )^{-1} \left(  \sum_i M_p \lambda_i  e_i^\top   \right) \left(  \sum_i  e_{i} \lambda_i^\top  M_p^\top \right) (2  X  X^\top )^{-1} X^\top \right]  \notag \\
& ~  - \sum_i  \lambda_i^\top  M_p^\top  (2 X^\top X )^{-1}    M_p \lambda_i  + \sum_i  \lambda_i^\top  W_{p}  e_i 
 \notag \\
= & ~   \tr \left[   \sum_i  X   (2 X^\top X )^{-1}  M_p \lambda_i   \lambda_i^\top M_p^\top  (2  X  X^\top )^{-1} X^\top \right]  \notag \\
& ~  - \sum_i  \lambda_i^\top  M_p^\top  (2  X^\top X )^{-1}    M_p \lambda_i + \sum_i  \lambda_i^\top  W_{p}  e_i 
 \notag \\
= & ~ \sum_i \lambda_i^\top  M_p^\top  (2 X  X^\top )^{-1}  X^\top  X   (2 X^\top X )^{-1}  M_p \lambda_i   \notag \\
& ~  - \sum_i  \lambda_i^\top  M_p^\top  (2 X^\top X )^{-1}    M_p \lambda_i + \sum_i \lambda_i^\top   W_{p}   e_i 
 \notag \\
= & ~  -  \frac{1}{2} \sum_i  \lambda_i^\top  M_p^\top  (2 X^\top  X )^{-1}  M_p \lambda_i  + \sum_i  \lambda_i^\top W_{p} e_i \notag \\
\end{align}

Note that $e_{i}^\top e_{i} =1 $ and $e_{i}^\top  e_{j} = 0$, if $i \neq j$.   Besides,  we can switch the  position of  two terms   in the trace function, such as $X   (2 X^\top  X )^{-1}     M_p \lambda_i$  and $ \lambda_i^\top M_p^\top  (2 X  X^\top )^{-1}  X^\top  $.    Further, after switching  the  two terms in the trace function, if the output is  a scale,  we can omit the trace function.

The gradients with reference to $  \lambda$ should be 0, \ie,
\begin{align}
\frac{\delta g(  \lambda)}{\delta   \lambda_i} = -    M_p^\top  (2   X^\top   X )^{-1}    M_p   \lambda_i  +   W_{p}   e_i =   0, 
\end{align}
We can obtain the optimal $  \lambda$ as below,
\begin{align}
  \lambda_i^* =  (M_p^\top   (2   X^\top   X )^{-1}   M_p  )^{-1}   W_{p}   e_i, 
\end{align}

The optimal $\delta W$ can be derived as below,
\begin{align}\label{eq:optimal_w}
\delta   W^*  = & ~ - (2   X   X^\top )^{-1}   \left(  \sum_i   M_p  (M_p^\top   (2   X^\top   X )^{-1}   M_p)^{-1}   W_{p}   e_i   e_i^\top   \right)  \notag  \\
=&~ - (2   X   X^\top )^{-1}   \left(  \sum_i   M_p   (M_p^\top   (2   X^\top   X )^{-1}   M_p  )^{-1}    M_p^\top   W   e_i   e_i^\top   \right).
\end{align} 

The minimal loss/error corresponding to the optimal $\delta   W$  can be obtained by
\begin{align}  \label{app:eq:optimal_loss}
L^* = & ~ \frac{1}{2}  \sum_i    \lambda_i^\top   M_p^\top   (2   X^\top   X )^{-1}   M_p   \lambda_i  \notag \\
= & ~ \frac{1}{2}  \sum_i   e_i^\top   W_{p}^\top (M_p^\top   (2   X^\top   X )^{-1}   M_p)^{-1}       M_p^\top   (2   X^\top   X )^{-1}   M_p (M_p^\top   (2   X^\top   X )^{-1}   M_p)^{-1}   W_{p}   e_i   \notag \\
= & ~ \frac{1}{2}  \sum_i   e_i^\top   W_{p}^\top  (M_p^\top   (2   X^\top   X )^{-1}   M_p)^{-1}   W_{p}   e_i   \notag \\
= & ~ \frac{1}{2}  \sum_i  (W^\top    M_p (M_p^\top   (2   X   X^\top )^{-1}   M_p)^{-1}    M_p^\top   W)_{i,i}.
\end{align}

\paragraph{Optimal Solution.} 
As demonstrated in Eq. \eqref{eq:optimal_w},   since $  e_{i}  $ is a one-hot vector, $   A     e_{i}   e_{i}^\top $  only has non-zero values in the $i$-th column with all zeros for all other columns.  Thus, in the sum of Eq.~\eqref{eq:optimal_w}, each term  indexed by $i$ just computes the $i$-th column of the output. Furthermore,  the computation of the $i$-th column does not affect the $j$-th column, $ \forall j \ne i$. For each column, we have the following, 
{\small
\begin{align}  
(\delta   W^*)_{*,i}
=  & ~ - (2   X   X^\top )^{-1}   \left(  M_p  (M_p^\top   (2   X^\top   X )^{-1}   M_p)^{-1}    M_p^\top   W   e_i   \right) ,  \notag \\
=  & ~ \left(- (2 X X^\top )^{-1}   \left(  M_p (M_p^\top   (2   X^\top   X )^{-1}   M_p)^{-1}    M_p^\top   W    \right)  \right)_{*,i}
\end{align} }%
Thus we can obtain the following optimal solution, 
\begin{align}\label{app:eq:optimal_w_final}
\delta   W^* 
=- (2   X   X^\top )^{-1}   \left(  M_p  ( M_p^\top   (2   X^\top   X )^{-1}   M_p)^{-1}    M_p^\top   W  \right) . 
\end{align}

\input{algorithms/general-algorithm}

%% file: tables/0.2-llama-1-results-appendix.tex
\begin{table*}[!ht]
\centering
\resizebox{0.9\linewidth}{!}{
\begin{tabular}{c|c|ccc|ccc|ccc|ccc}
\toprule
\multirow{2}{*}{Method} & \multirow{2}{*}{\begin{tabular}[c]{@{}c@{}}Prune\\ Ratio\end{tabular}} & \multicolumn{3}{|c|}{LLaMA-7B} & \multicolumn{3}{|c|}{LLaMA-13B} & \multicolumn{3}{|c|}{LLaMA-30B} & \multicolumn{3}{|c}{LLaMA-65B} \\
                        &                                                                        & Wiki     & PTB     & C4      & Wiki     & PTB      & C4      & Wiki     & PTB      & C4      & Wiki     & PTB      & C4      \\ \midrule
Baseline                & \textbackslash{}                                                       & 5.68            & 27.34           & 7.08            &     5.09           &      19.23           &   6.61              &       4.10         &        16.29         &        5.98        &    3.53           &    17.61            &      5.62          \\ \midrule
LLM-Pruner              & 10\%                                                                   & 7.41            & 36.73           & 9.25            & 6.38           & 31.85           & 8.16            & 4.92           & 18.17           & 6.63           & 3.98          & 19.44          & 6.08           \\
SliceGPT                & 10\%                                                                   & 6.97            & 88.48           & 23.54           & 6.11           & 60.15           & 20.18           & 5.24           & 39.72           & 17.83          & 4.57          & 36.20          & 14.14          \\
FLAP                    & 10\%                                                                   & 6.34            & 32.39           & 8.058           & 5.45           & 20.99           & 7.33            & 4.52           & 17.29           & 6.49           & 3.91          & 19.35          & 6.04           \\
Ours                    & 10\%                                                                   & \textbf{6.01}   & \textbf{31.65}  & \textbf{7.94}   & \textbf{5.38}  & \textbf{20.52}  & \textbf{7.27}   & \textbf{4.43}  & \textbf{17.26}  & \textbf{6.47}  & \textbf{3.82} & \textbf{19.28} & \textbf{6.02}  \\ \midrule
LLM-Pruner              & 20\%                                                                   & 10.73           & 59.73           & 12.15           & 6.38           & 31.85           & 9.42            & 5.83           & 20.18           & 7.55           & 4.65          & 21.85          & 6.75           \\
SliceGPT                & 20\%                                                                   & 8.42            & 120.89          & 35.93           & 7.17           & 86.26           & 29.70           & 6.18           & 50.95           & 26.85          & 5.34          & 61.09          & 21.86          \\
FLAP                    & 20\%                                                                   & 7.40            & 36.77           & 9.99            & 6.03           & 23.33           & 8.42            & 5.18           & 19.30           & 7.42           & 4.45          & 21.45          & 6.75           \\
Ours                    & 20\%                                                                   & \textbf{6.60}   & \textbf{35.75}  & \textbf{9.49}   & \textbf{5.89}  & \textbf{23.11}  & \textbf{8.39}   & \textbf{4.92}  & \textbf{18.58}  & \textbf{7.36}  & \textbf{4.26} & \textbf{20.94} & \textbf{6.73}  \\ \midrule
LLM-Pruner              & 30\%                                                                   & 18.58           & 93.24           & 17.78           & 11.81          & 45.42           & 12.65           & 7.59           & 24.97           & 9.08           & 5.52          & 26.38          & 7.53           \\
SliceGPT                & 30\%                                                                   & 12.75           & 258.90          & 67.33           & 9.18           & 125.40          & 46.46           & 7.74           & 75.89           & 42.71          & 6.56          & 74.43          & 35.68          \\
FLAP                    & 30\%                                                                   & 9.18            & 47.35           & 13.08           & 6.97           & 27.36           & 10.01           & 6.28           & 21.88           & 8.53           & 5.10          & 23.91          & 7.59           \\
Ours                    & 30\%                                                                   & \textbf{7.56}   & \textbf{41.05}  & \textbf{11.53}  & \textbf{6.57}  & \textbf{26.27}  & \textbf{9.98}   & \textbf{5.46}  & \textbf{20.48}  & \textbf{8.46}  & \textbf{4.75} & \textbf{22.13} & \textbf{7.51}  \\ \midrule
LLM-Pruner              & 40\%                                                                   & 38.27           & 238.09          & 29.97           & 20.24          & 75.17           & 18.81           & 10.59          & 40.47           & 11.76          & 6.92          & 31.46          & 9.03           \\
SliceGPT                & 40\%                                                                   & 250.2          & 1657         & 834.8          & 13.80          & 196.9          & 83.46           & 11.53          & 136.3          & 78.77          & 9.11          & 115.2         & 64.33          \\
FLAP                    & 40\%                                                                   & 12.34           & 65.54           & 16.95           & 8.67           & 35.91           & 12.31           & 8.73           & 29.11           & 10.12          & 5.83          & 23.61          & 8.55           \\
Ours                    & 40\%                                                                   & \textbf{8.95}   & \textbf{52.84}  & \textbf{14.56}  & \textbf{7.41}  & \textbf{29.54}  & \textbf{12.01}  & \textbf{6.21}  & \textbf{22.90}  & \textbf{9.93}  & \textbf{5.33} & \textbf{22.79} & \textbf{8.42}  \\ \midrule
LLM-Pruner              & 50\%                                                                   & 126.0          & 460.7          & 73.88           & 45.69          & 152.99          & 36.94           & 19.68          & 78.29           & 18.64          & 9.34          & 43.79          & 12.16          \\
SliceGPT                & 50\%                                                                   & 1540         & 6364         & 4847         & 18.75          & 277.34          & 122.5          & 15.60          & 195.4          & 118.5         & 12.01         & 160.3         & 92.66          \\
FLAP                    & 50\%                                                                   & 21.89           & 135.8          & 30.86           & 12.88          & 53.54           & 18.37           & 13.41          & 47.30           & 13.17          & 6.98          & 28.52          & 10.36          \\
Ours                    & 50\%                                                                   & \textbf{11.66}  & \textbf{82.55}  & \textbf{20.72}  & \textbf{8.91}  & \textbf{37.56}  & \textbf{16.12}  & \textbf{7.25}  & \textbf{26.68}  & \textbf{11.91} & \textbf{6.02} & \textbf{25.17} & \textbf{9.73}  \\ \midrule
LLM-Pruner              & 60\%                                                                   & 2194         & 1164         & 1327         & 357.5         & 588.7          & 189.1          & 71.71          & 288.1          & 55.66          & 24.91         & 168.1         & 35.05          \\
SliceGPT                & 60\%                                                                   & 2559         & 6401         & 5816         & 31.92          & 459.4          & 223.2          & 28.40          & 338.3          & 208.9         & 20.84         & 278.3         & 168.9         \\
FLAP                    & 60\%                                                                   & 57.51           & 626.2          & 120.4          & 31.36          & 206.7          & 42.99           & 12.84          & 54.01           & 18.96          & 9.79          & 58.73          & 17.66          \\
Ours                    & 60\%                                                                   & \textbf{18.91}  & \textbf{195.15} & \textbf{40.68}  & \textbf{12.15} & \textbf{68.04}  & \textbf{26.18}  & \textbf{8.78}  & \textbf{34.30}  & \textbf{16.02} & \textbf{7.10} & \textbf{32.53} & \textbf{12.43} \\ \midrule
LLM-Pruner              & 70\%                                                                   & 9010         & 4111         & 2655         & 5900        & 6039         & 1334         & 895.7         & 3274         & 456.9         & Nan           & Nan            & Nan            \\
SliceGPT                & 70\%                                                                   & 3605         & 7304         & 8096         & 67.65          & 874.9          & 537.4          & 71.25          & 633.1          & 406.6         & 102.4        & 863.9         & 662.8         \\
FLAP                    & 70\%                                                                   & 577.9          & 1835         & 833.7          & 647.8         & 1588         & 975.1          & 2786        & 2735         & 2416        & Nan           & 2333        & Nan            \\
Ours                    & 70\%                                                                   & \textbf{162.9} & \textbf{721.3} & \textbf{361.6} & \textbf{41.66} & \textbf{275.7} & \textbf{115.3} & \textbf{39.88} & \textbf{124.2} & \textbf{50.43} & \textbf{9.65} & \textbf{69.49} & \textbf{20.84} \\
\bottomrule
\end{tabular}
}
\caption{
Perplexity ($\downarrow$) results for LLaMA-1 family models with different pruning ratios on WikiText2, PTB, and C4 with 2048 sequence length.
}
\label{tab:main_llama_1_results_appendix}
\end{table*}

%% file: tables/3-llama_2-results.tex
\begin{table*}[!ht]
\centering
\resizebox{0.77\linewidth}{!}{
\begin{tabular}{c|c|ccc|ccc|ccc}
\toprule
\multirow{2}{*}{Method} & \multirow{2}{*}{\begin{tabular}[c]{@{}c@{}}Prune\\ Ratio\end{tabular}} & \multicolumn{3}{c|}{LLaMA-2-7B}                   & \multicolumn{3}{c|}{LLaMA-2-13B}                  & \multicolumn{3}{c}{LLaMA-2-70B}                 \\
                        &                                                                        & Wiki           & PTB            & C4             & Wiki           & PTB            & C4             & Wiki          & PTB            & C4             \\
                        \midrule
Baseline                & \textbackslash{}                                                       & 5.47           & 22.51          & 6.97           & 4.88           & 28.87          & 6.47           & 3.32          & 15.65          & 5.52           \\
\midrule
LLM-Pruner              & 10\%                                                                   & 6.70           & 30.60          & 8.65           & 5.94           & 34.46          & 7.61           & 3.79          & 16.78          & 5.91           \\
SliceGPT                & 10\%                                                                   & 6.86           & 63.25          & 24.66          & 6.01           & 69.44          & 21.81          & 4.44          & 37.37          & 15.09          \\
FLAP                    & 10\%                                                                   & 6.04           & 25.52          & 8.04           & 5.33           & 35.17          & 7.40           & 3.62          & 16.67          & 5.90           \\
Ours                    & 10\%                                                                   & \textbf{5.88}  & \textbf{25.39} & \textbf{7.98}  & \textbf{5.23}  & \textbf{34.14} & \textbf{7.33}  & \textbf{3.52} & \textbf{16.26} & \textbf{5.88}  \\
\midrule
LLM-Pruner              & 20\%                                                                   & 9.35           & 42.27          & 11.53          & 8.27           & 54.67          & 9.77           & 4.18          & 18.97          & 6.82           \\
SliceGPT                & 20\%                                                                   & 8.38           & 103.5          & 37.50          & 7.26           & 101.1          & 33.08          & 5.28          & 51.92          & 23.19          \\
FLAP                    & 20\%                                                                   & 7.12           & 31.70          & 10.18          & 6.09           & 43.63          & 9.27           & 4.15          & 18.75          & 6.72           \\
Ours                    & 20\%                                                                   & \textbf{6.76}  & \textbf{31.60} & \textbf{10.10} & \textbf{5.93}  & \textbf{43.59} & \textbf{9.14}  & \textbf{3.90} & \textbf{18.67} & \textbf{6.69}  \\
\midrule
LLM-Pruner              & 30\%                                                                   & 16.18          & 74.09          & 17.09          & 12.01          & 98.56          & 13.37          & 4.97          & 20.97          & 7.59           \\
SliceGPT                & 30\%                                                                   & 11.64          & 172.78         & 60.43          & 9.84           & 166.0          & 54.05          & 6.67          & 79.59          & 38.31          \\
FLAP                    & 30\%                                                                   & 8.92           & 42.85          & 13.05          & 7.14           & 51.69          & 11.57          & 4.84          & 20.63          & 7.56           \\
Ours                    & 30\%                                                                   & \textbf{7.92}  & \textbf{42.28} & \textbf{12.98} & \textbf{6.95}  & \textbf{49.71} & \textbf{11.50} & \textbf{4.43} & \textbf{20.32} & \textbf{7.55}  \\
\midrule
LLM-Pruner              & 40\%                                                                   & 42.22          & 175.1          & 34.32          & 21.03          & 178.7          & 20.51          & 7.01          & 26.60          & 9.74           \\
SliceGPT                & 40\%                                                                   & 18.92          & 295.0          & 108.2          & 15.74          & 241.35         & 96.82          & 9.83          & 134.12         & 73.26          \\
FLAP                    & 40\%                                                                   & 13.02          & 70.23          & 19.34          & 9.10           & 70.46          & 14.96          & 5.86          & 23.05          & 8.84           \\
Ours                    & 40\%                                                                   & \textbf{9.82}  & \textbf{61.52} & \textbf{17.97} & \textbf{8.10}  & \textbf{61.29} & \textbf{14.34} & \textbf{5.11} & \textbf{22.58} & \textbf{8.77}  \\
\midrule
LLM-Pruner              & 50\%                                                                   & 165.1          & 375.3          & 149.0          & 41.71          & 308.5          & 43.42          & 9.95          & 41.62          & 13.96          \\
SliceGPT                & 50\%                                                                   & 25.94          & 391.8          & 151.0          & 21.27          & 319.66         & 133.08         & 13.20         & 186.35         & 109.7          \\
FLAP                    & 50\%                                                                   & 25.14          & 190.5          & 39.87          & 14.39          & 126.8          & 22.36          & 7.09          & 28.62          & 10.58          \\
Ours                    & 50\%                                                                   & \textbf{12.92} & \textbf{100.6} & \textbf{27.03} & \textbf{9.71}  & \textbf{89.20} & \textbf{19.26} & \textbf{5.91} & \textbf{27.78} & \textbf{10.51} \\
\midrule
LLM-Pruner              & 60\%                                                                   & Nan            & 5389           & Nan            & 169.9          & 803.0          & 157.5          & 35.72         & 139.7          & 48.60          \\
SliceGPT                & 60\%                                                                   & 43.95          & 602.7          & 305.0          & 35.52          & 488.7          & 214.9          & 22.87         & 317.2          & 192.9          \\
FLAP                    & 60\%                                                                   & 58.60          & 271.4          & 150.5          & 31.32          & 559.4          & 52.21          & 9.33          & 40.29          & 14.64          \\
Ours                    & 60\%                                                                   & \textbf{19.73} & \textbf{249.6} & \textbf{53.35} & \textbf{13.19} & \textbf{203.3} & \textbf{32.44} & \textbf{7.06} & \textbf{36.22} & \textbf{13.88} \\
\bottomrule
\end{tabular}
}
\caption{
Results of LLaMA-2 model family.
}
\label{tab:llama_2_results}
\end{table*}

%% file: tables/4-llama_3-results.tex
\begin{table*}[!ht]
\centering
\resizebox{0.53\linewidth}{!}{
\begin{tabular}{c|c|ccc|ccc}
\toprule
\multirow{2}{*}{Method} & \multirow{2}{*}{\begin{tabular}[c]{@{}c@{}}Prune\\ Ratio\end{tabular}} & \multicolumn{3}{c|}{LLaMA-3-8B}                   & \multicolumn{3}{c}{LLaMA-3-70B}                 \\
                        &                                                                        & Wiki           & PTB            & C4             & Wiki          & PTB            & C4             \\
                        \midrule
Baseline                & \textbackslash{}                                                       & 6.14           & 10.60          & 8.92           & 2.86          & 8.17           & 6.76           \\
\midrule
SliceGPT                & 10\%                                                                   & 10.55          & 70.80          & 72.01          & 6.93          & 82.40          & 45.76          \\
FLAP                    & 10\%                                                                   & 7.40           & 12.98          & 11.89          & 3.86          & 8.87           & 7.97           \\
Ours                    & 10\%                                                                   & \textbf{7.05}  & \textbf{12.81} & \textbf{11.81} & \textbf{3.65} & \textbf{8.74}  & \textbf{7.92}  \\
\midrule
SliceGPT                & 20\%                                                                   & 15.61          & 147.5          & 122.9          & 8.57          & 143.9          & 73.17          \\
FLAP                    & 20\%                                                                   & 9.30           & 16.26          & 16.82          & 5.49          & 9.82           & 9.57           \\
Ours                    & 20\%                                                                   & \textbf{8.58}  & \textbf{16.23} & \textbf{16.69} & \textbf{5.15} & \textbf{9.51}  & \textbf{9.53}  \\
\midrule
SliceGPT                & 30\%                                                                   & 29.68          & 255.9          & 212.2          & 12.22         & 231.6          & 123.2          \\
FLAP                    & 30\%                                                                   & 12.70          & 22.35          & 23.37          & 6.82          & 10.74          & 11.87          \\
Ours                    & 30\%                                                                   & \textbf{10.97} & \textbf{21.41} & \textbf{23.07} & \textbf{6.39} & \textbf{10.53} & \textbf{11.11} \\
\midrule
SliceGPT                & 40\%                                                                   & 54.94          & 472.8          & 334.0          & 23.35         & 393.5          & 263.0          \\
FLAP                    & 40\%                                                                   & 19.04          & 37.86          & 37.61          & 8.23          & 12.37          & 12.96          \\
Ours                    & 40\%                                                                   & \textbf{13.95} & \textbf{28.53} & \textbf{31.07} & \textbf{7.52} & \textbf{12.00} & \textbf{12.94} \\
\midrule
SliceGPT                & 50\%                                                                   & 91.28          & 615.2          & 410.47         & 37.82         & 553.0          & 360.6          \\
FLAP                    & 50\%                                                                   & 39.44          & 80.21          & 85.60          & 10.21         & 15.72          & 15.87          \\
Ours                    & 50\%                                                                   & \textbf{18.51} & \textbf{53.01} & \textbf{48.36} & \textbf{8.82} & \textbf{14.45} & \textbf{15.69} \\
\bottomrule
\end{tabular}
}
\caption{
Results of LLaMA-3 model family.
}
\label{tab:llama_3_results}
\end{table*}

%% file: tables/6-mmlu-gsm8k-llama_1-llama_2.tex

\begin{table*}[!ht]
\centering
\resizebox{0.87\linewidth}{!}{
\begin{tabular}{c|cc|cc|cc|cc|cc}
\toprule
Sparsity & \multicolumn{2}{c|}{LLaMA-7B} & \multicolumn{2}{c|}{LLaMA-13B} & \multicolumn{2}{c|}{LLaMA-30B} & \multicolumn{2}{c|}{LLaMA-2-7B} & \multicolumn{2}{c}{LLaMA-2-13B} \\
\midrule
Dataset  & MMLU         & GSM8K         & MMLU          & GSM8K         & MMLU          & GSM8K         & MMLU          & GSM8K          & MMLU           & GSM8K          \\
\midrule
Dense    & 34.9         & 9.7           & 46.9          & 16.9          & 58.2          & 36.3          & 45.8          & 15.4           & 54.9           & 23.6           \\
\midrule
10\%     & 31.8         & 4.3           & 40.5          & 11.0          & 53.1          & 17.9          & 39.3          & 9.2            & 48.3           & 15.6           \\
20\%     & 29.3         & 2.3           & 36.3          & 6.5           & 47.7          & 13.3          & 33.1          & 6.5            & 41.0           & 9.5            \\
30\%     & 27.4         & 1.5           & 33.8          & 3.4           & 42.5          & 6.5           & 30.1          & 3.1            & 36.1           & 3.9            \\
40\%     & 26.5         & 2.7           & 29.7          & 2.0           & 37.6          & 4.9           & 27.4          & 1.5            & 28.8           & 1.8            \\
50\%     & 25.2         & 1.5           & 29.1          & 2.3           & 35.3          & 1.6           & 25.6          & 1.6            & 27.7           & 1.6        \\
\bottomrule
\end{tabular}
}
\caption{
Results of LLaMA model family on MMLU and GSM8K datasets.
}
\label{tab:mmlu_gsm8k}
\end{table*}

%% file: tables/1-zeroshot-results-appendix.tex
\begin{table*}[!ht]
\centering
\resizebox{0.9\linewidth}{!}{
\begin{tabular}{c|c|ccccccc|c}

\toprule
Method & Prune Ratio   & BoolQ & PIQA  & Hella Swag & Wino Grande & ARC-e  & ARC-c  & OBQA  & Average Acc.         \\
\toprule
LLaMA-7B  & /          & 73.18 & 78.35 & 72.99           & 67.01      & 67.45          & 41.38       & 42.40 & 63.25           \\
\toprule
LLM-Pruner(v)     & \multirow{3}{*}{10\%}          & 67.95 & 77.42 & 69.31           & 63.54      & 66.33          & 39.85       & 41.20 & 60.80          \\
LLM-Pruner(e2)   &           & 68.29 & 76.88 & 70.25           & 64.33      & 65.28          & 40.10        & 39.60 & 60.68          \\
LLM-Pruner(e1)   &           & 66.97 & 77.26 & 70.30            & 64.33      & 65.24          & 40.19       & 41.00   & 60.76          \\
                            \midrule
SliceGPT         &   10\%       & 57.68 & 69.80  & 59.32           & 68.11      & 62.75          & 36.01       & 38.00   & 55.95           \\
                            \midrule
FLAP             &   10\%         & 74.43 & 75.41 & 68.68           & 67.01      & 65.78          & 38.48       & 41.00   & 61.54          \\
\midrule
Ours          &   10\%    & 71.07 & 77.53 & 71.75 & 68.11          & 61.45      & 39.59          & 41.6        & \textbf{61.58} \\
\toprule
LLM-Pruner(v)         & \multirow{3}{*}{20\%}         & 61.44 & 71.71 & 57.27           & 54.22      & 55.77          & 33.96       & 38.40 & 53.25           \\
LLM-Pruner(e2)        &         & 59.39 & 75.57 & 65.34           & 61.33      & 59.18          & 37.12       & 39.80 & 56.82           \\
LLM-Pruner(e1)       &           & 57.06 & 75.68 & 66.80            & 59.83      & 60.94          & 36.52       & 40.00   & 56.69           \\
                            \midrule
SliceGPT      & 20\%          & 37.89 & 64.09 & 45.67           & 62.75      & 53.62          & 31.74       & 33.20 & 46.99          \\
                            \midrule
FLAP           & 20\%            & 68.59 & 74.21 & 64.98           & 64.40       & 59.89          & 37.80        & 40.20 & 58.58           \\
\midrule
Ours            & 20\%     & 67.92 & 74.76 & 67.31 & 66.54 & 58.80 & 36.77 & 39.40  & \textbf{58.79}  \\
\bottomrule
\toprule
LLaMA-13B                & /           & 68.47 & 78.89 & 76.24           & 70.09      & 74.58          & 44.54       & 42.00   & 64.97           \\
\toprule
LLM-Pruner(c)     & \multirow{2}{*}{10\%}          & 68.47 & 74.76 & 66.99           & 66.38      & 66.58          & 35.24       & 38.20 & 59.52           \\
LLM-Pruner(b)    &           & 70.64 & 78.40  & 75.00           & 69.46      & 72.82 & 41.47       & 41.40 & 64.17           \\
                            \midrule
SliceGPT     & 10\%             & 61.74 & 69.97 & 60.74           & 69.38      & 66.79          & 40.70        & 41.80 & 58.73           \\
                            \midrule
FLAP            & 10\%         & 63.76 & 78.07 & 73.69           & 69.61      & 69.53          & 39.93       & 41.60 & 62.31           \\
\midrule
Ours            & 10\%      & 70.21 & 77.97 & 76.04 & 69.69 & 70.92 & 43.00 & 41.80 & \textbf{64.23}  \\
\toprule
LLM-Pruner(c)    & \multirow{2}{*}{20\%}           & 62.39 & 66.87 & 49.17           & 58.96      & 49.62          & 31.83       & 33.20 & 50.29           \\
LLM-Pruner(b)    &             & 67.68 & 77.15 & 73.41           & 65.11      & 68.35          & 38.40        & 42.40 & 61.79           \\
                            \midrule
SliceGPT         & 20\%           & 50.34 & 66.00    & 53.37           & 68.11      & 60.56          & 36.35       & 38.20 & 53.27           \\
                            \midrule
FLAP       & 20\%          & 62.23 & 76.50  & 70.59           & 68.35      & 65.66          & 38.99       & 41.60 & 60.56           \\
\midrule
Ours       & 20\%      & 63.18 & 77.04 & 71.81 & 70.40 & 66.79 & 41.55 & 42.40 & \textbf{61.88} \\

\bottomrule

\end{tabular}
}
\caption{
Pruning results for LLaMA-7B and LLaMA-13B on common sense reasoning datasets.
LLM-Pruner (v), (e2), and (e1) denote vector-wise and element-wise, while (c) and (b) represent channel and block pruning.
}
\label{tab:zeroshot_results_full_appendix}
\end{table*}

%% file: tables/2-128-length-main-results-appendix.tex

\begin{table*}[!ht]
\centering
\resizebox{0.97\linewidth}{!}{
\begin{tabular}{c|ccc|ccc|ccc|ccc|ccc}
\toprule
Prune Ratio & \multicolumn{3}{c|}{10\%} & \multicolumn{3}{c|}{20\%} & \multicolumn{3}{c|}{30\%} & \multicolumn{3}{c|}{40\%} & \multicolumn{3}{c}{50\%}  \\
\midrule
Dataset     & Wiki   & PTB    & C4     & Wiki   & PTB    & C4     & Wiki   & PTB    & C4     & Wiki   & PTB    & C4     & Wiki   & PTB    & C4    \\
LLM-Pruner  & 15.37  & 64.54  & 12.27  & 19.09  & 82.03  & 15.51  & 30.64  & 116.9  & 21.29  & 52.28  & 212.6  & 33.69  & 122.8  & 419.7  & 67.67  \\
SliceGPT    & 14.52  & 113.2  & 33.68  & 19.27  & 154.1  & 58.23  & 44.96  & 349.1  & 164.5  & 535.5  & 2005   & 1505   & 2241   & 7401   & 5224 \\
FLAP        & 13.84  & 62.28  & 11.51  & 14.62  & 67.17  & 13.95  & 17.62  & 78.30  & 17.71  & 22.32  & 100.8  & 22.22  & 31.80  & 157.4  & 34.51   \\
Ours        & \textbf{13.31} & \textbf{61.16} & \textbf{11.45} & \textbf{14.47} & \textbf{65.96} & \textbf{14.20} & \textbf{16.40} & \textbf{74.52} & \textbf{17.52} & \textbf{19.04} & \textbf{89.62} & \textbf{21.69} & \textbf{23.32} & \textbf{125.9} & \textbf{29.87}  \\
\bottomrule
\end{tabular}
}
\caption{
Pruning results for LLaMA-7B model on WikiText2, PTB, and C4 with 128 sequence length.
}
\label{tab:128_sequence_length_full_appendix}
\end{table*}

%% file: sections/8-preliminary.tex
\section{Preliminary}\label{sec:preliminary}
In this section, we present preliminary concepts, several basic facts, and definitions for our paper. 

\subsection{Notations}\label{sec:preli:notations}
For two vectors $x \in \mathbb{R}^n$ and $y \in \mathbb{R}^n$, we use $\langle x, y \rangle$ to denote the inner product between $x,y$, i.e., $\langle x, y\rangle = \sum_{i=1}^n x_i y_i$.
We use $e_i$ to denote a vector where only $i$-th coordinate is $1$, and other entries are $0$.
For each $a, b \in \mathbb{R}^n$, we use $a \circ b \in \mathbb{R}^n$ to denote the vector where $i$-th entry is $(a\circ b)_i = a_i b_i$ for all $i \in [n]$.
We use ${\bf 1}_n$ to denote a length-$n$ vector where all the entries are ones.
We denote the $i$-th row of a matrix $A$ as $A_i$. 
We use $x_{i,j}$ to denote the $j$-th coordinate of $x_i \in \mathbb{R}^n$.
We use $\|x\|_p$ to denote the $\ell_p$ norm of a vector $x \in \mathbb{R}^n$, i.e. $\|x\|_1 := \sum_{i=1}^n |x_i|$, $\|x\|_2 := (\sum_{i=1}^n x_i^2)^{1/2}$, and $\|x\|_{\infty} := \max_{i \in [n]} |x_i|$.
For $k > n$, for any matrix $A \in \mathbb{R}^{k \times n}$, we use $\|A\|$ to denote the spectral norm of $A$, i.e. $\|A\|:=\sup_{x\in \mathbb{R}^n} \|Ax\|_2 / \|x\|_2$.
For a tensor $X \in \R^{B \times N \times D}$ and a matrix $U \in \R^{D \times d_1}$, we define $Y = X \cdot U \in \R^{B \times N \times d_1}$.
For a matrix $V \in \R^{d_2 \times B}$ and a tensor $X \in \R^{B \times N \times D}$, we define $Z =  V \cdot X \in \R^{d_2 \times N \times D}$.
For a square matrix $A$, we use $\tr[A]$ to denote the trace of $A$, \ie, $\tr[A] = \sum_{i=1}^n A_{i,i}$. 

\subsection{Facts}
Here we provide facts we use.
\begin{fact}[Norm bounds]\label{fac:bound}
For $a, b \in \R^{d}$ and $A \in \R^{n \times d}$, we have
\begin{itemize}
    \item $\|a \circ b\|_2 \leq \|a \|_2 \cdot \|b \|_2$.
    \item $\|A b\|_2 \leq \|A\| \cdot \|b\|_2$.
\end{itemize}
\end{fact}

\begin{fact}[Linear algebra]\label{fac:linalg}
For $a, b \in \R^{d}$ and $A, B \in \R^{n \times d}$ and $C \in \R^{d \times d}$, we have
\begin{itemize}
    \item $a \circ b = b \circ a = \diag(a)b = \diag(b)a$.
    \item $\diag(a) C \diag(b) = (a b^\top) \circ C$.
    \item $ A B^\top  = \sum_{i \in [d]} A_{*,i} B_{*,i}^\top$. 
\end{itemize}
\end{fact}

\begin{fact}[Calculus]\label{fac:calculus}
For $a, b \in \R^{d}$ and $A \in \R^{n \times d}$, we have
\begin{itemize}
    \item $\frac{\d a \circ b}{\d a_j} = b_j e_j$ for $j \in [d]$. 
    \item $\frac{\d \langle a, b \rangle}{\d a_j} = b_j$ for $j \in [d]$.
    \item $\frac{\d A b}{\d b} = A$.
    \item $\frac{\d 0.5 \|b\|_2^2}{\d a} = a^\top \frac{\d b}{\d a}$.
\end{itemize}
\end{fact}

\subsection{Internal Computation Alignment}

Assume the input $X \in \mathbb{R}^{B\times N\times D}$ where $B$ denotes the batch size, $N$ denotes the number of tokens, $D$ denotes the dimension of hidden states, and the pruned weights can be denoted as $W_q, W_k, W_v \in \mathbb{R}^{D \times D_p} $ and $W_o \in \mathbb{R}^{D_p \times D}$ where $D_p \leq D$ denotes the pruned dimension size.
The computation in the self-attention mechanism can be presented as follows,
\begin{align*}
    \mathrm{Attention}(Q, K, V) := \mathrm{Softmax} ( QK^{\top} / \sqrt{D_k}  ) \cdot V \ \in \mathbb{R}^{B \times N \times D_p},
\end{align*}
where $Q  = X \cdot W_q, K  = X \cdot W_k, V  = X \cdot W_v$.
Multiplying $W_o$, we have the final output: 
\begin{align*} 
    Z^\attn = \mathrm{Attention}(Q, K, V) \cdot W_o\ \in \mathbb{R}^{B \times N \times D}.
\end{align*} 

Similarly, for the MLP module, we define the up, gate, and down projection weights as $W_{\mathrm{up}}, W_{\mathrm{gate}} \in \mathbb{R}^{D \times D_p}$ and $W_{\mathrm{down}} \in \mathbb{R}^{D_p \times D}$, respectively. 
With the same input $X \in \mathbb{R}^{B \times N \times D}$, the MLP module can be expressed as follows,
\begin{align*}
    H_\mathrm{Up} :=  X \cdot W_{\mathrm{up}} \in \mathbb{R}^{B \times N \times D_{p}} ~~\text{and}~~ H_\mathrm{Gate} := X \cdot W_{\mathrm{gate}}  \in \mathbb{R}^{B \times N \times D_{p}}.
\end{align*}
The final output then is
\begin{align*}
Z^\mlp =  (H_\mathrm{Up} \circ H_\mathrm{Gate}) \cdot W_{\mathrm{down}} \in \mathbb{R}^{B \times N \times D}.
\end{align*}


%% file: sections/9-theory.tex
\newpage
\clearpage

\section{Theory}

\subsection{More Related Work}

\paragraph{Optimization.}
Optimization is a cornerstone of computer science and mathematics, encompassing various techniques to find optimal solutions. 
Linear Programming (LP) and Semi-Definite Programming (SDP) form a fundamental basis in this field, addressing problems with linear/quadratic objective functions and constraints, offering powerful tools for diverse applications \cite{a00,dgj+06,aw08,dkk+16,dhl19,cls19,jlt20,jkl+20,gs22,syz23}. 
The concept of dynamic maintenance has gained prominence, focusing on efficient solution updates as input data changes \cite{cls19,lsz19,b20,jlsw20,blss20,jswz21,sy21,dly21,b21,jkl+20,hjs+22,gs22}.  
In machine learning, optimization techniques have significantly impacted Support Vector Machines (SVMs) \cite{cl01,j06,gsz23,gswy23,tlto23,bsz23,lsz23, gms23} and Empirical Risk Minimization (ERM) \cite{n83,v91,pj92,bbm05,bb08,njls09,mb11,fgrw12,n13,jz13,v13,sz13,db14,fgks15,zx17,jlgj18,lsz19,lszz23,bsy23}, enhancing their efficiency and performance. 
Recent work continues to expand the application of optimization across various domains, including graph algorithms, numerical methods, and advanced machine learning architectures \cite{ll18,dzp+18,adh+19a,adh+19b,sy19,cgh+19,zmg19,cg19,zg19,os20,jt19,lss+20,hls+21,zpd+20,bps+20,zks+20,szz21,als+22,zha22,gms23,lsz23,qsy23,csy24,ssx23_nns,qss23_gnn,dms23_spar,lls+24,llsz24_nn_tw}, demonstrating its ongoing significance in advancing computational capabilities.

\paragraph{Attention Theory.}
Attention mechanisms have been extensively studied and developed over the years, becoming a fundamental component in various neural network architectures. 
\cite{dls23,lsx+23,gll+24a} investigate the softmax regression problem, while \cite{syz23} propose and analyze the attention kernel regression problem. 
The rescaled hyperbolic functions regression is examined by~\cite{gsy23}.
Following this, \cite{swy23,lswy23} delve into two-layer attention regression problems. 
\cite{gll+24b} demonstrate that attention layers in transformers learn two-dimensional cosine functions. 
Additionally, \cite{dsxy23} explore data recovery using attention weights, and~\cite{kmz23,sxy23} investigate the replacement of the softmax unit with a polynomial unit.
Moreover, some works theoretically explore variations or combinations of the attention mechanism with other techniques, such as quantum attention~\cite{gsyz23_quantum, gsyz24_quantum_kro}, tensor attention~\cite{as24_iclr, lssz24_tat}, and differentially private attention~\cite{gls+24_dp_tree, gsy23_dp, gls+24_dp_ntk} and other applications such as \cite{cklm19,tdp19,hl19,vb19,b22,bsz23,cls+24,cll+24_icl,cll+24_rope,lss+24_relu,lls+24_io,lss+24_mutlilayer,gls+24_diffusion,lssz24_tat}.

\subsection{Error Bound for Masked Weight}\label{sec:theory:error_bound}
In this section, we show how to derive the error bound for masked weight. We assume $B = 1$ for simplicity of proofs.
\begin{lemma}[formal version of Lemma~\ref{lem:XW_upper_bound:informal}]\label{lem:XW_upper_bound:formal}
If the following conditions hold:
\begin{itemize}
    \item Let $W \in \R^{D \times D'}$, $X \in \R^{N \times D}$.
    \item Let $M \in \{0,1\}^{D}$ and $\rho \in [0,1]$ be ratio of number of zeros to number of entries in $M$.
    \item Assume $\|X\| \leq R$.
\end{itemize}
We can show that for $i \in [D']$ we have
\begin{align*}
    \|X W_{*,i} - X(M \circ W_{*,i})\|_2 \leq \rho R \| W_{*,i}\|_2
\end{align*}
\end{lemma}
\begin{proof}
We can show
\begin{align*}
    \|XW_{*,i} - X(M \circ W_{*,i})\|_2 = & ~ \|X( W_{*,i}- M \circ W_{*,i}))\|_2 \\
    \leq & ~ \|X\| \cdot \|({\bf 1}_D-M) \circ W_{*,i}\|_2 \\
    \leq & ~ \|X\| \cdot \|({\bf 1}_D-M)\|_2 \cdot \| W_{*,i}\|_2\\
    \leq & ~ \rho R \| W_{*,i}\|_2
\end{align*}
where the first step follows from basic algebra, the second step and the third step follow from Fact~\ref{fac:bound}, the last step follows from we assume $\|X\| \leq R$ and the definition of $\rho$.
\end{proof}

\subsection{Find Optimal Mask}\label{sec:theory:optimal_mask}
In this section, we show how to use Newton's method to find the numerical score defined in Section~\ref{sec:numerical_score}.
\begin{theorem}[Mask optimization, formal version of Theorem~\ref{thm:mask_optimize:informal}]\label{thm:mask_optimize:formal}
If the following conditions hold:
\begin{itemize}
    \item Let $W \in \R^{D \times D'}$, $X \in \R^{N \times D}$.
    \item Let $z \in [0,1]^{D}$.
    \item Let $r \in [0,D]$ denote the number of ones (it can be a fractional number). 
    \item Let $\lambda >0 $ denote a regularization co-efficients.
    \item Assume $\|X\| \leq R$. 
\end{itemize}
There exist an algorithm (Algorithm~\ref{alg:main}) that can get the optimal $z$ such that
\begin{align*}
    \underset{z \in [0,1]^D }{\arg \min} & ~~~~ \frac{1}{2}\sum_{i \in [D']} \|X W_{*,i} - X(z \circ W_{*,i})\|_2^2 + \frac{1}{2}\lambda \cdot ( \langle {\bf 1}_D , z \rangle - r )^2
\end{align*}
\end{theorem}
\begin{proof}
We can use Newton's method to solve this problem. 
To use Newton's method we need to compute the gradient and Hessian.
After calculation in Lemma~\ref{lem:g_loss}, \ref{lem:g_reg}, \ref{lem:H_loss}, and \ref{lem:H_reg}, we use Algorithm~\ref{alg:main} to get optimal $z$.
\end{proof}

\subsubsection{Gradient Calculation}
In this section, we compute the gradient for $L(z)$ and $L_{\mathrm{reg}}(z)$. First, we define $f(z)$ for simplicity to calculate the gradient.
\begin{definition}\label{def:fz}
We define $f(z)_i := X(z \circ W_{*,i}) \in \R^D$ for $i \in [D']$.
\end{definition}
Then, we calculate the gradient for each loss.
\begin{lemma}[Gradient of loss function]\label{lem:g_loss}
If the following conditions hold
\begin{itemize}
    \item Let $L(z)_i  :=  0.5 \|X W_{*,i} - X(z \circ W_{*,i})\|_2^2$ for $i \in [D']$.
\end{itemize}
Then, we can show
\begin{itemize}
\item Part 1. For each $i\in [D'], j \in [D]$
    \begin{align*}
        \underbrace{\frac{\d f(z)_i}{\d z_j}}_{\mathrm{D}} = X_{*,j} W_{j,i} 
    \end{align*}
\item Part 2. For each $i\in [D'],j \in [D]$
\begin{align*}
    \underbrace{ \frac{\d L(z)_i }{ \d z_j} }_{ \mathrm{scalar} } = (X(z \circ W_{*,i}) - X W_{*,i})^\top X_{*,j} W_{j,i}
\end{align*}
\item Part 3. For each $i \in [D']$
\begin{align*}
    \underbrace{ \frac{\d L(z)_i }{ \d z} }_{ D } = ((W_{*,i} W_{*,i}^\top ) \circ (X^\top X)) (z - {\bf 1}_D) 
\end{align*}
\item Part 4. 
\begin{align*}
    \underbrace{ \frac{\d L(z)}{ \d z} }_{ D } = ((W W^\top) \circ (X^\top X)) (z - {\bf 1}_D)
\end{align*}
\end{itemize}
\end{lemma}

\begin{proof}
{\bf Proof of part 1.}
We can show 
\begin{align*}
    \frac{\d f(z)_i}{\d z_j} = & ~ \frac{\d X(z \circ W_{*,i})}{\d z_j}\\
    = & ~ X(e_j W_{j,i})\\
    = & ~ X_{*,j} W_{j,i} 
\end{align*}
where the first step follows from Definition~\ref{def:fz}, the second step follows from Fact~\ref{fac:calculus}, and the last step follows from simple algebra.

{\bf Proof of part 2.}
We can show 
\begin{align*}
    \frac{\d L(z)_i }{ \d z_j} = & ~ \frac{\d 0.5 \|X W_{*,i} - X(z \circ W_{*,i})\|_2^2 }{ \d z_j}\\
    = & ~ (XW_{*,i} - f(z)_i)^\top \frac{\d (XW_{*,i} - f(z)_i)}{\d z_j}\\
    = & ~ (f(z)_i - XW_{*,i})^\top \frac{\d f(z)_i}{\d z_j}\\
    = & ~ (X(z \circ W_{*,i}) - X W_{*,i})^\top X_{*,j} W_{j,i}
\end{align*}
where the first step follows from definition of $L(z)_i$, the second step follows from Fact~\ref{fac:calculus}, the third step follows from simple algebra, and the last step follows from {\bf Part 1}.

{\bf Proof of part 3.}
We can show
\begin{align*}
    \frac{\d L(z)_i }{ \d z} 
    = & ~ ((X(z \circ W_{*,i}) - X W_{*,i})^\top X \diag(W_{*,i}))^\top \\
    = & ~ \diag(W_{*,i}) X^\top (X(z \circ W_{*,i}) - X W_{*,i})\\
    = & ~ \diag(W_{*,i}) X^\top X (W_{*,i} \circ (z - {\bf 1}_D))\\
    = & ~ \diag(W_{*,i}) X^\top X \diag(W_{*,i})(z - {\bf 1}_D)\\
    = & ~ ((W_{*,i} W_{*,i}^\top ) \circ (X^\top X)) (z - {\bf 1}_D) 
\end{align*}
where the first step follows from {\bf Part 2}, the second step follows from property of transpose, the third step follows from simple algebra, the fourth step follows from Fact~\ref{fac:linalg}, and the last step follows from Fact~\ref{fac:linalg}.

{\bf Proof of part 4.}
We can show
\begin{align*}
    \frac{\d L(z) }{ \d z} = & ~ \frac{\d }{\d z} \sum_{i \in [D']} L(z)_i \\
    = & ~ \sum_{i \in [D']} \frac{\d L(z)_i}{\d z}  \\
    = & ~ \sum_{i \in [D']} ((W_{*,i} W_{*,i}^\top ) \circ (X^\top X)) (z - {\bf 1}_D) \\
    = & ~ ((W W^\top) \circ (X^\top X)) (z - {\bf 1}_D) 
\end{align*}
where the first step follows from definition of $L(z)$, the second step follows from basic calculus, the third step follows from {\bf Part 3}, and the last step follows from Fact~\ref{fac:linalg}. 
\end{proof}

\begin{lemma}[Gradient of regularization]\label{lem:g_reg}
If the following conditions hold
\begin{itemize}
    \item Let $L_{\mathrm{reg}}(z)  := 0.5 \lambda \cdot ( \langle {\bf 1}_D , z \rangle - r)^2$.
\end{itemize}
Then, we can show
\begin{itemize}
\item Part 1. For each $j \in [D]$
\begin{align*}
    \underbrace{ \frac{\d L_{\mathrm{reg}}(z) }{ \d z_j} }_{ \mathrm{scalar} } = \lambda (\langle {\bf 1}_D , z \rangle - r ) 
\end{align*}
\item Part 2.
\begin{align*}
    \underbrace{ \frac{\d L_{\mathrm{reg}}(z) }{ \d z} }_{ D } = \lambda (\langle {\bf 1}_D , z \rangle - r )  \cdot {\bf 1}_D
\end{align*}
\end{itemize}
\end{lemma}

\begin{proof}
{\bf Proof of part 1.}
We can show
\begin{align*}
    \frac{\d L_{\mathrm{reg}}(z) }{ \d z_j} = & ~ \frac{\d  0.5 \lambda \cdot ( \langle {\bf 1}_D , z \rangle - r )^2}{ \d z_j} \\
    = & ~ \lambda (\langle {\bf 1}_D , z \rangle - r ) \frac{\d \langle {\bf 1}_D , z \rangle - r }{\d z_j}\\
    = & ~ \lambda (\langle {\bf 1}_D , z \rangle - r ) 
\end{align*}
where the first step follows from definition of $L_{\mathrm{reg}}(z)$, the second step follows from basic calculus, and the last step follows from Fact~\ref{fac:calculus}.

{\bf Proof of part 2.}
Using {Part 1}, we can show
\begin{align*}
    \frac{\d L_{\mathrm{reg}}(z) }{ \d z} = \lambda (\langle {\bf 1}_D , z \rangle - r )  \cdot {\bf 1}_D
\end{align*}
\end{proof}

\subsubsection{Hessian Calculation}
In this section, we calculate the hessian for $L(z)$ and $L_{\mathrm{reg}}(z)$.
\begin{lemma}[Hessian of loss function]\label{lem:H_loss}
If the following conditions hold
\begin{itemize}
    \item Let $L(z)_i  :=  0.5 \|X W_{*,i} - X(z \circ W_{*,i})\|_2^2$ for $i \in [D']$.
\end{itemize}
Then, we can show
\begin{align*}
    \underbrace{ \frac{\d^2 L(z)}{ \d z^2} }_{ D \times D} = (W W^\top) \circ (X^\top X)
\end{align*}
\end{lemma}

\begin{proof}
We can show
\begin{align*}
     \frac{\d^2 L(z)}{ \d z^2} 
     = & ~ \frac{\d }{\d z} \frac{\d L(z)}{ \d z}\\
     = & ~ \frac{\d }{\d z} ((W W^\top) \circ (X^\top X)) (z - {\bf 1}_D) \\
     = & ~ (W W^\top) \circ (X^\top X)
\end{align*}
where the first step follows from basic calculus, the second step follows from Lemma~\ref{lem:g_loss}, and the last step follows from Fact~\ref{fac:calculus}.
\end{proof}

\begin{lemma}[Hessian of regularization]\label{lem:H_reg}
If the following conditions hold
\begin{itemize}
    \item Let $L_{\mathrm{reg}}(z)  := 0.5 \lambda \cdot ( \langle {\bf 1}_D , z \rangle - r)^2$.
\end{itemize}
Then, we can show 
\begin{align*}
    \underbrace{ \frac{\d^2 L_{\mathrm{reg}}(z) }{ \d z^2} }_{ D \times D} = \lambda {\bf 1}_{D \times D}
\end{align*}
\end{lemma}

\begin{proof}
We can show 
\begin{align*}
    \frac{\d^2 L_{\mathrm{reg}}(z) }{ \d z^2} = & ~ \frac{\d }{\d z} \frac{\d L_{\mathrm{reg}}(z) }{ \d z}\\
    = & ~ \frac{\d }{\d z} \lambda (\langle {\bf 1}_D , z \rangle - r )  \cdot {\bf 1}_D\\
    = & ~ \frac{\d }{\d z} \lambda \langle {\bf 1}_D , z \rangle \cdot {\bf 1}_D \\
    = & ~ \lambda {\bf 1}_{D \times D}
\end{align*}
where the first step follows from basic calculus, the second step follows from Lemma~\ref{lem:g_reg}, the third step follows from basic calculus, and the last step follows from basic calculus.
\end{proof}